\documentclass[lettersize,journal]{IEEEtran}
\usepackage{amsmath,amsfonts}
\usepackage{algorithmic}
\usepackage{array}
\usepackage{subfigure}
\usepackage{textcomp}
\usepackage{stfloats}
\usepackage{url}
\usepackage{verbatim}
\usepackage{graphicx}
\usepackage{cite}

\usepackage{mathrsfs}
\usepackage{amssymb}
\usepackage{multirow}
\usepackage[ruled]{algorithm2e}
\usepackage{float}
\usepackage{xcolor}
\usepackage[colorlinks=true,linkcolor=black,citecolor=blue,urlcolor=blue,]{hyperref}
\begin{document}

\title{Deep Reinforcement Learning-Based User Scheduling for Collaborative Perception}

\author{Yandi Liu, Guowei Liu,~\IEEEmembership{Graduate Student Member,~IEEE}, Le Liang,~\IEEEmembership{Member,~IEEE}, \\ Hao Ye,~\IEEEmembership{Member,~IEEE}, Chongtao Guo,~\IEEEmembership{Member,~IEEE}, Shi Jin,~\IEEEmembership{Fellow,~IEEE}
\thanks{Yandi Liu, Guowei Liu, Le Liang and Shi Jin are with the National Mobile Communications Research Laboratory, Southeast University, Nanjing 210096, China (e-mail: \{yandi\_liu, grownliu, lliang, jinshi\}@seu.edu.cn). Le Liang is also with the Purple Mountain Laboratories, Nanjing 211111, China.

     Hao Ye is with the Department of Electrical and Computer Engineering, University of California, Santa Cruz, CA 95064,
     USA (e-mail: yehao@ucsc.edu).

     Chongtao Guo is with the College of Electronics and Information
     Engineering, Shenzhen University, Shenzhen 518060, China (e-mail: ctguo@
     szu.edu.cn).
}% <-this % stops a space
}

% The paper headers
\markboth{Journal of \LaTeX\ Class Files,~Vol.~14, No.~8, January~2025}%
{Shell \MakeLowercase{\textit{et al.}}: A Sample Article Using IEEEtran.cls for IEEE Journals}

\maketitle

\begin{abstract}
Stand-alone perception systems in autonomous driving suffer from limited sensing ranges and occlusions at extended distances, potentially resulting in catastrophic outcomes. To address this issue, collaborative perception is envisioned to improve perceptual accuracy by using vehicle-to-everything (V2X) communication to enable collaboration among connected and autonomous vehicles and roadside units. However, due to limited communication resources, it is impractical for all units to transmit sensing data such as point clouds or high-definition video. As a result, it is essential to optimize the scheduling of communication links to ensure efficient spectrum utilization for the exchange of perceptual data. In this work, we propose a deep reinforcement learning-based V2X user scheduling algorithm for collaborative perception. Given the challenges in acquiring perceptual labels, we reformulate the conventional label-dependent objective into a label-free goal, based on characteristics of 3D object detection. Incorporating both channel state information (CSI) and semantic information, we develop a double deep Q-Network (DDQN)-based user scheduling framework for collaborative perception, named SchedCP. Simulation results verify the effectiveness and robustness of SchedCP compared with traditional V2X scheduling methods. Finally, we present a case study to illustrate how our proposed algorithm adaptively modifies the scheduling decisions by taking both instantaneous CSI and perceptual semantics into account.
\end{abstract}


\section{Introduction}
\IEEEPARstart{A}{utonomous} driving has garnered significant attention in recent years for its capability of transforming intelligent transportation systems. To improve safety, the perception module plays a pivotal role in comprehending the driving environment and provides important information for the planning and control modules \cite{chen2022milestones}. However, stand-alone perception systems suffer from restricted sensing ranges and occlusions at extended distances, which may lead to disastrous consequences \cite{liu2023towards}. To address this challenge, collaborative perception has been developed to improve perceptual ability by using vehicular communication to enable information sharing among different vehicles and roadside units (RSUs) \cite{ren2022collaborative}. A typical illustration of collaborative perception is depicted in Fig. 1, where the ego vehicle is unable to sense the presence of the object on the left due to occlusion, and it receives perceptual information from the connected and autonomous vehicle (CAV) to obtain the position of this occluded object, thereby allowing it to adjust driving decisions to avoid collision.

\begin{figure}[t]\label{cp_intro}
\centering
\includegraphics[width=0.87\linewidth]{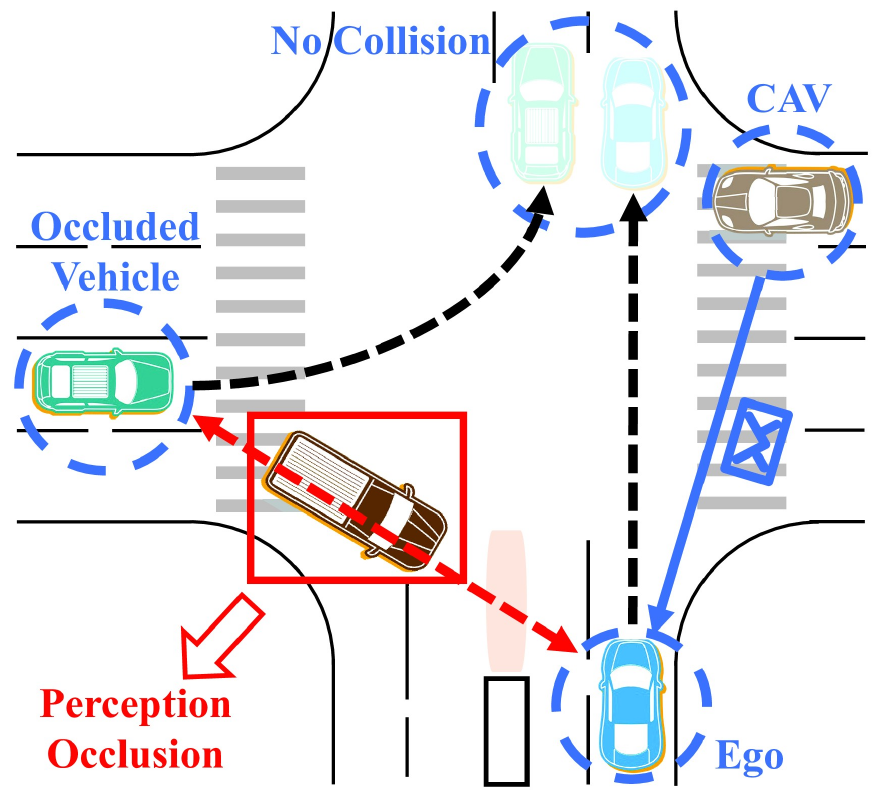}
\caption{An illustrative example of collaborative perception.}
\end{figure}

The transition from stand-alone perception to collaborative perception is facilitated by vehicular communications. Vehicular communications, also known as vehicle-to-everything (V2X) communications, significantly enhance road safety and traffic efficiency by enabling real-time data exchange among vehicles, RSUs, and pedestrians, etc., thereby facilitating adaptive decision-making of transportation systems \cite{liang2017vehicular}. Given its importance, the 3rd generation partnership project (3GPP) has developed standards on vehicular communications, including LTE-V2X \cite{3rd2016technical} and NR-V2X \cite{3rd2019technical}, to support its application in LTE and 5G networks. With its rapid development, V2X technologies have been pivotal in collaborative perception systems, as they improve vehicular perception and responsiveness to dynamic road conditions, leading to a reduction in traffic accidents and congestion \cite{wang2018networking}. However, early research efforts on collaborative perception mainly focused on the design of information fusion mechanisms \cite{li2021learning,xu2022v2x,wang2020v2vnet} and transmitted feature selection\cite{hu2022where2comm} while neglecting the design in terms of V2X communication. These studies generally assume flawless communication between CAVs and RSU, which may make these methods difficult to implement in practical applications. In practice, the maximum aggregated data rate for V2X unicast is approximately 50 Mbps per link \cite{3rd2024technical}. Nonetheless, the real-time transmission of light detection and ranging (LiDAR) point clouds demands link rate greater than 100 Mbps \cite{feng2020real}. Consequently, it is challenging for V2X networks to accommodate the transmission of all raw sensor data or perceptual features.

Given the limited vehicular network communication resources, it is imperative to develop effective resource allocation techniques for vehicular communication. Recent advancements in resource allocation techniques for vehicular networks can be divided into two categories: optimization-based and learning-based approaches. In the optimization-based category, the resource allocation problem is usually modeled as an optimization problem with constraints. Traditional optimization techniques are then employed to adjust the allocation of available resources, such as resource blocks and transmit power, to maximize the associated utility value in a given scenario \cite{bai2010low,sun2015cluster,liang2017resource}. However, these optimization-based methods may fail to work due to the complexity of the communication environment in reality and the difficulty in modeling V2X requirements mathematically. In contrast, learning-based approaches can dynamically adjust resource allocation under various communication environments since the implicit channel dynamics as well as quality-of-service (QoS) requirements for different links can be learned during the training phase. For instance, in \cite{ye2019deep}, taking both unicast and broadcast scenarios into consideration, a resource allocation framework for V2V communications based on single-agent reinforcement learning (SARL) was proposed. Along this line of thought, the application of multi-agent reinforcement learning to the spectrum sharing problem in vehicular networks was discussed in \cite{liang2019spectrum,huang2023meta,nasir2019multi}, enabling multiple links to achieve effective collaborative resource allocation. However, these resource allocation methods are not specifically designed for autonomous driving tasks such as collaborative perception, therefore disregarding the impact of perception semantics which are actually crucial for cooperative perception \cite{qin2024ai,sheng2024semantic}. This may result in suboptimal perception performance despite achieving satisfactory communication performance.

In response, there have been a few studies that have delved into the V2X user scheduling problem for collaborative perception. Specifically, in \cite{qiu2021autocast}, a system was proposed to enable scalable, infrastructure-less collaborative perception through direct V2V communication based on visibility and relevance. A mobility-aware sensor scheduling strategy (MASS) was developed in \cite{jia2023mass}, which leveraged the predictable aspects of vehicular mobility. On the basis of \cite{jia2023mass}, a second-order approximation of the perception topology was used in the C-MASS system in \cite{jia2024c}. In \cite{liu2024select2col}, the Select2Col framework focused on optimizing bandwidth usage by leveraging the spatial-temporal importance of semantic information. In \cite{ye2023accuracy}, the problem of joint subtask placement and resource allocation in per-object granularity was formulated, which was then solved by an iterative method based on the genetic algorithm.

\begin{figure*}[htbp]
\centering
\includegraphics[width=0.86\linewidth]{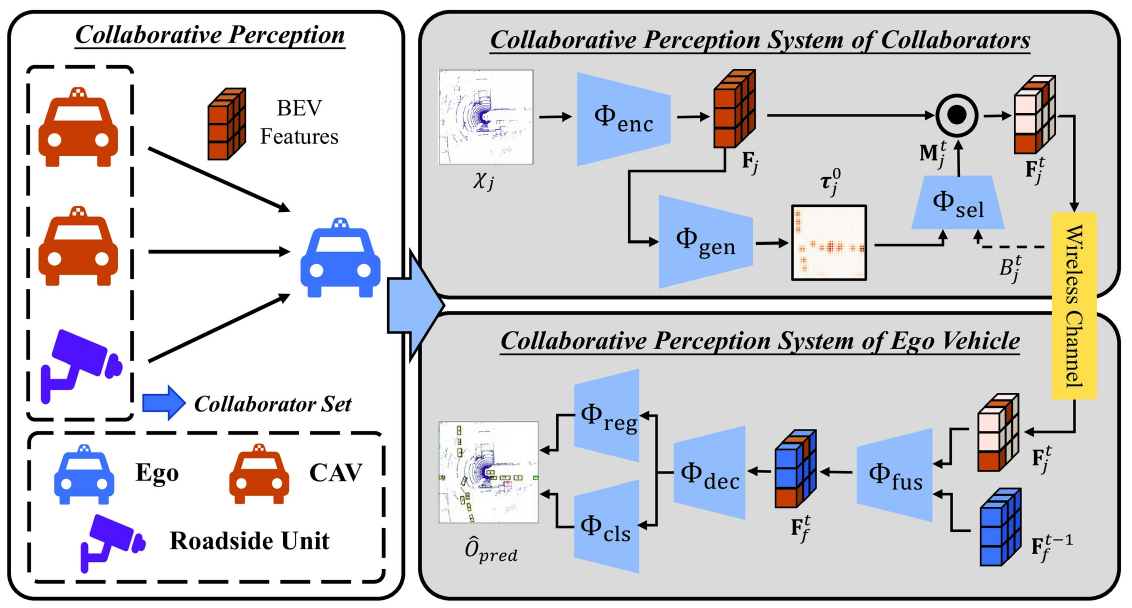}
\caption{An illustration of the collaborative perception framework for autonomous driving.}
\end{figure*}

Despite these advancements, existing works on user scheduling for collaborative perception still face notable limitations. On one hand, these works consider scheduling intervals on the order of hundreds of milliseconds, roughly as the sensor sampling interval, and disregard small-scale fading. Nevertheless, the impact of small-scale fading should be considered in the resource allocation design due to the high dynamics of vehicular communication networks. On the other hand, there are some flaws in the design of the objective function for the optimization problem or the loss function for the training of deep learning models in these works. For example, the Select2Col framework in \cite{liu2024select2col} requires perceptual labels for supervised training, which introduces significant overhead for annotating the perceptual targets. Although different label-free utility functions have been devised in \cite{qiu2021autocast,jia2023mass,jia2024c,ye2023accuracy} to characterize the perception gain due to collaboration between different units, they are all based on the final detection results and thus difficult to portray the real-time perceptual gain during the collaboration process.

Motivated by the above background, we explicitly model the problem of V2X user scheduling with the consideration of small-scale fading and design a DRL-based user scheduling framework for collaborative perception, learning the pattern of semantic and channel changes in the collaboration process to dynamically adjust user scheduling decisions. Our main contributions are summarized as follows:
\begin{enumerate}
    \item We mathematically formulate the joint optimization problem of bird's eye view (BEV) feature selection and V2X user scheduling in V2X-aided collaborative perception systems. Given the difficulty of obtaining labels and the computational complexity of the optimization objective, we transform the label-dependent objective into a heuristic label-free objective.
    \item We propose a novel V2X user scheduling framework for collaborative perception systems based on double deep Q-Network (DDQN), named SchedCP. Considering the inherent limitations in label availability, we design both label-dependent and label-free rewards during the training phase, intended for flexible training patterns and perception scenarios.
    \item We conduct extensive simulation on the V2X-Sim collaborative perception dataset under different communication settings to validate the effectiveness and robustness of the algorithm compared with several baselines. We also design a case study to explain its mechanism and advantages.
\end{enumerate}

The remainder of this paper is organized as follows. We introduce the system model and problem formulation in Section II. In Section III, we present the detailed framework of DDQN-based V2X user scheduling for collaborative perception. Simulation results and case study are provided in Section IV, and the conclusions are drawn in Section V.

\section{System Model And Problem Formulation}

\subsection{System Model}
As shown in Fig. 2, we consider a collaborative perception system at an intersection with $N + 1$ units, forming the set $\mathcal{M}$ where there exists an ego vehicle that receives perceptual information. The first vehicle in the set denotes the ego vehicle. The remaining $N$ units, including CAVs or RSUs, constitute the set of collaborators for the ego vehicle, denoted by $\mathcal{N}$. There is a scheduler on the ego side that decides which links (from the collaborators) will be activated to access the spectrum. The scheduled collaborators transmit their BEV features to the ego vehicle through wireless channels. The ego vehicle receives the BEV features from scheduled collaborators, fuses them with its local feature, and utilizes the fused BEV feature for 3D object detection.

The LiDAR point clouds collected by the $j$-th collaborator and the ego vehicle can be denoted as $\chi_j, j \in \mathcal{N}$ and $\chi_e$, respectively. To share perceptual information from multiple collaborators, the ego vehicle broadcasts its pose information to all collaborators. Since the data collected by LiDAR of each vehicle or RSU is in its own coordinate system, it is necessary to transform the point cloud of all collaborators to the ego vehicle's coordinate system. Each collaborator, based on its own pose information and the pose information of the ego vehicle, aligns the collected point clouds to the coordinate system of the ego vehicle, denoted as $\chi_{j \rightarrow e}$. All collaborators and the ego vehicle then encode the transformed point cloud $\chi_{j \rightarrow e}$ through a feature encoder $\Phi_{\text {enc}}$ to obtain the BEV features $\mathbf{F}_j \in \mathbb{R}^{H \times W \times D}$, where $H \times W$ is the spatial dimension. This process can be denoted by
\begin{equation}\label{bbb}
\mathbf{F}_j=\Phi_{\text {enc}}\left(\chi_{j \rightarrow e}\right), j \in \mathcal{M},
\end{equation}
where $\Phi_{\text {enc}}$ follows the architecture of Pointpillars \cite{lang2019pointpillars} and $\mathbf{F}_j\left(x,y\right) \in \mathbb{R}^{D}$ represents the feature of the grid $\left(x,y\right)$, where each grid $\left(x,y\right)$ is a coordinate in the spatial dimension.

The BEV features of all units are subsequently processed through their confidence generation network $\Phi_{\text {gen}}$ to obtain the initial spatial confidence map $\boldsymbol{\boldsymbol{\tau}}_j^0 \in \mathbb{R}^{H \times W}$ at the beginning of the sensor sampling interval, represented by
\begin{equation}\label{confmap_gen}
\boldsymbol{\tau}^0_j=\Phi_{\text {gen}}\left(\mathbf{F}_j\right), j \in \mathcal{M}.
\end{equation}
This will be utilized for feature selection to be introduced later in Section III. To effectively manage the transmission of perceptual data, the sensor sampling interval can be divided into $T$ scheduling slots. This means that there are $T$ scheduling decisions occurring in one sensor sampling interval. In the $t$-th scheduling slot, each collaborator selects part of all local BEV features, denoted as $\mathbf{F}_{j}^t \in \mathbb{R}^{H \times W \times D}$, and transmits them to the ego vehicle according to the spatial confidence map, the number of allowable transmitted grids $B_j^t$, and a pre-defined BEV feature selection rule, represented by
\begin{equation}\label{Mask}
\mathbf{M}_{j}^t=\Phi_{\text {sel}}\left(\boldsymbol{\tau}^0_j,  B_j^t\right), j \in \mathcal{N}, t \in \mathcal{T},
\end{equation}
\begin{equation}\label{bbb}
\mathbf{F}_{j}^t=\mathbf{F}_j \odot \mathbf{M}_{j}^t, j \in \mathcal{N}, t \in \mathcal{T},
\end{equation}
where $\mathcal{T} = \{1, \ldots, T\}$, $\odot$ is the element-wise multiplication, and $\Phi_{\text {sel}}$ represents the feature selection network. $\mathbf{M}_{j}^t \in \{0,1\}^{H \times W}$ stands for the feature selection mask of the $j$-th collaborator at the $t$-th scheduling slot, where $\mathbf{M}_{j}^t\left(x,y\right) = 1$ represents that $\mathbf{F}_j\left(x,y\right)$ is chosen for transmission and $\mathbf{M}_{j}^t\left(x,y\right) = 0$ otherwise. The number of allowable transmitted grids $B_j^t$ is determined by the data rate of the scheduled link and the size of the feature in each grid, which will be discussed in detail later. It is noteworthy that all networks mentioned above are the same for $\forall j \in \mathcal{N}$, allowing for the omission of the corresponding subscripts.

The ego vehicle gathers BEV features from every CAV through a V2V link and communicates with the RSU through a V2I link. The $j$-th link represents the link between the $j$-th collaborator and the ego vehicle. Due to the highly dynamic nature of vehicular networks, the channels of V2V or V2I links also change within a scheduling slot. Based on this, we set that each scheduling slot consists of $T_s$ sub-time slots as depicted in Fig. 3, where there are channel variations between adjacent sub-time slots within each scheduling slot. The power gain $g_j^{t, t_s}$ of the $j$-th link in the $t_s$-th sub-time slot of the $t$-th scheduling slot is represented as
\begin{equation}\label{power gain}
g_j^{t, t_s}=\alpha_j\left|h_j^{t, t_s}\right|^2, j \in \mathcal{N},
\end{equation}
where $\alpha_j$ accounts for the large-scale fading, including both path loss and shadowing, which remains constant throughout the whole sensor sampling interval. The small-scale fading $h_j^{t, t_s}$ changes at each sub-time slot, following a first-order Markov process \cite{liang2017spectrum}, denoted by
\begin{equation}\label{fast_fading}
h_j^{t, t_s}=\mu_j h_j^{t, t_s-1}+e, j \in \mathcal{N},
\end{equation}
where $\mu_j=J_0\left(2 \pi v_j f_c \Delta t / c\right)$ characterizes the correlation between the small-scale fading of adjacent sub-time slots. $J_0\left(\cdot\right)$ is the zero-order Bessel function, $v_j$ is the relative velocity of the $j$-th collaborator with respect to the ego vehicle, $f_c$ is the carrier frequency, $\Delta t$ is the duration of sub-time slot, $c$ is the speed of light, and $e \sim \mathcal{CN}\left(0,1-\mu^2\right)$ portrays the uncertainty in the variation of small-scale fading between neighboring sub-time slots.

\begin{figure}[t]
\centering
\includegraphics[width=\linewidth]{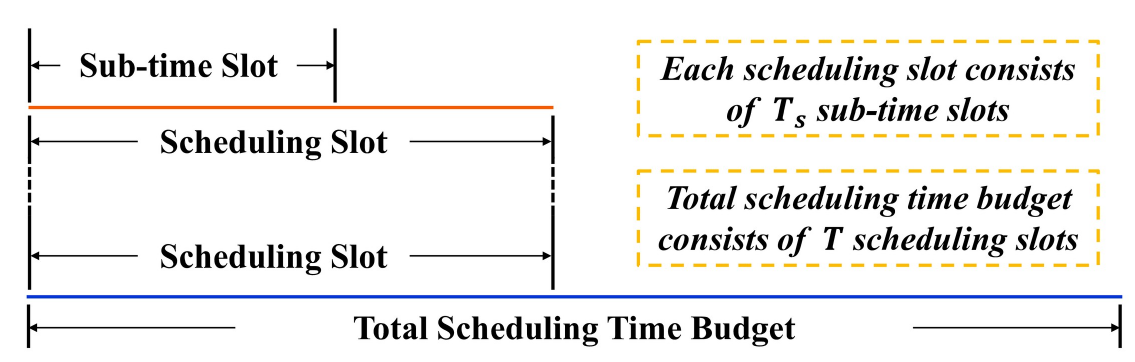}
\caption{Relationship among the sensor sampling interval, scheduling slot and sub-time slot.}
\end{figure}

We can derive the expression for the number of allowable transmitted grids in the feature $\mathbf{F}_j$, i.e. $B_j^t$ in (\ref{Mask}), in the $t$-th scheduling slot as
\begin{equation}\label{bbb}
B_j^t=\sum_{t_s=1}^{T_s} \frac{C_j^{t, t_s}\Delta t}{D} = \sum_{t_s=1}^{T_s} \frac{W}{D} \Delta t \log \left(1+\frac{P g_i^{t, t_s}}{N_0 W}\right),
\end{equation}
where $C_j^{t, t_s}$ is the instantaneous transmission rate at the $t_s$-th sub-time slot in the $t$-th scheduling slot and $D$ is the BEV feature data size of each grid. $P$ is the transmit power, $W$ is the bandwidth, $N_0$ is the power spectral density of the additive white Gaussian noise, and $\sigma^2 = N_0 W$ is the noise power.

For the ego vehicle, a scheduling decision must be made in each scheduling slot to determine which V2I link or V2V link can access the spectrum, which is represented as $\lambda_j^t \in\{0,1\}, j \in \mathcal{N}, t \in \mathcal{T}$. In the $t$-th scheduling slot, $\lambda_j^t = 1$ represents the $j$-th link's access to the spectrum, and otherwise $\lambda_j^t = 0$. For simplicity, we assume that the ego vehicle will activate only one link and channel fading is approximately the same within the whole spectrum band. We also assume these links share the same spectrum, thus effectively degenerating into a user scheduling problem in multiple access channels. In addition, we do not differentiate between V2V and V2I links in this study since CAVs and the RSU have similar functionalities and communication requirements for this work. Based on the scheduling decision, the ego vehicle integrates the BEV features sent by those links that have been granted access to the spectrum and passes them through the fusion network $\Phi_{\text {fus}}$ to obtain the fused feature $\mathbf{F}_f^t \in \mathbb{R}^{H \times W \times D}$ at the end of $t$-th scheduling slot, represented by
\begin{equation}\label{bbb}
\mathbf{F}_f^t=\Phi_{\text {fus}}\left(\mathbf{F}_f^{t-1},\left\{\mathbf{F}_{j }^t\right\}_{\lambda_j^t=1, j \in \mathcal{N}}\right),
\end{equation}
where $\mathbf{F}_f^0$ represents the local BEV feature of the ego vehicle without collaboration.

Upon completion of the scheduling decisions across all $T$ scheduling slots, the ego vehicle obtains the final fused BEV feature $\mathbf{F}_f^T$ of the frame at the end of the sensor sampling interval, and the fused feature $\mathbf{F}_f^T$ is then decoded and passed through the prediction network to obtain the final detection results as
\begin{equation}\label{bbb}
\hat{O}_{\text {pred}}^T=\Phi_{\text {pred}}\left(\Phi_{\text {dec}}\left(\mathbf{F}_f^T\right)\right),
\end{equation}
where $\hat{O}_{\text {pred}}^T$ stands for the final detection result at the end of the $T$-th scheduling slot and $\Phi_{\text {dec}}$ is the decoding network. $\Phi_{\text {pred}}$ represents the prediction network and it consists of the regression header $\Phi_{\text {reg}}$ and classification header $\Phi_{\text {cls}}$, where $\Phi_{\text {reg}}$ is responsible for predicting the three-dimensional position, dimensions, and yaw angle of the bounding boxes while $\Phi_{\text {cls}}$ generates the classification output $\mathbf{\hat{O}}_{\text {cls}} \in \mathbb{R}^{H \times W}$ that assigns a confidence score to each bounding box, which reflects the likelihood that it contains an object.

\subsection{Scheduling Problem Formulation}
The ultimate goal of this optimization problem is to maximize the perceptual accuracy of the ego, which can be defined by the average precision (AP) performance metric \cite{zamanakos2021comprehensive}. Two thresholds of 0.5 and 0.7 of Intersection of Union (IoU) are utilized to determine whether the detection is correct or not, i.e., the detection is judged to be correct when the IoU between the prediction box and the label exceeds the respective threshold. Specifically, the optimization aims to enhance both the BEV feature selection decision $\mathbf{M}_{j }^t$ and the V2X user scheduling decision at the ego $\lambda_j^t$ to optimize the perception performance of the ego vehicle. Therefore the optimization problem can be represented as
\begin{subequations}\label{prob1}
    \begin{align}
        \mathop {\max }\limits_{\left\{\mathbf{M}_{j}^t, \lambda_j^t\right\}_{j \in \mathcal{N}}^{t \in \mathcal{T}}}  \quad &\text{AP}\left(\hat{O}_{\text {pred}}^T, O_{\text {true}}\right) \label{obj1}
        \\
        \ \textrm{ s.t. } \qquad
        & \lambda_j^t \in\{0,1\}, \forall j \in \mathcal{N}, t \in \mathcal{T}, \label{st3} \\
        & \sum_{j \in \mathcal{N}} \lambda_j^t \leq 1, \forall t \in \mathcal{T}, \label{st1} \\
        & \sum_{(x, y)} \mathbf{M}_{j}^t(x, y) \leq B_j^t, \forall j \in \mathcal{N}, \label{st2}
    \end{align}
\end{subequations}
where $O_{\text {true}}$ stands for the label for the detection results and $\text{AP}(\hat{O}_{\text {pred}}^T, O_{\text {true}})$ represents the AP metric calculated by the prediction $\hat{O}_{\text {pred}}^T$ and the label $O_{\text {true}}$. In this problem formulation, (\ref{st1}) constrains the maximum number of links accessing the spectrum, and (\ref{st2}) specifies that the transmission rate cannot exceed the channel capacity to ensure distortion-free transmission of the BEV features.

\section{SchedCP: DRL-Based V2X User Scheduling Framework for Collaborative Perception}

In this section, we delve into the details of the SchedCP V2X user scheduling framework. We first introduce the feature selection method based on the spatial confidence map, and then we design the user scheduling algorithm based on DDQN. This algorithm is capable of adapting the decision-making of link accessing the spectrum according to the instantaneous changes of channel state information (CSI) and perceptual semantic information. 

\subsection{Feature Selection Based on Spatial Confidence Map}\label{FS}
In this subsection, we propose a feature selection method based on the spatial confidence map. For spatial confidence map $\boldsymbol{\boldsymbol{\tau}} \in \mathbb{R}^{H \times W}$, each grid $(x, y)$ in the map represents the probability that the corresponding region is occupied by an object. A comparison between Fig. 4(a) and (b) reveals that the regions covered by the location of the final detected targets have a higher value for the corresponding grids in the spatial confidence map. Denoting the spatial confidence map as $\boldsymbol{\tau}_e^t$ for the ego vehicle and  $\boldsymbol{\tau}_j^t$ for the $j$-th collaborator at the beginning of $t$-th scheduling slot, their transition can be expressed as
\begin{equation}
\centering
\begin{aligned}\label{conf_p}
& \left\{\begin{array}{lc}
\boldsymbol{\tau}_e^t = \Phi_{\text {gen}}\left(\mathbf{F}_f^{t}\right), \\
\boldsymbol{\tau}_j^t  = \boldsymbol{\tau}_j^{t-1} \odot \left(1 - \mathbf{M}_{j }^{t-1}\right), j \in \mathcal{N}.
\end{array}\right. \\
\end{aligned}
\end{equation}
In other words, the ego vehicle regenerates a new spatial confidence map based on the new fused feature $\mathbf{F}_f^{t}$ at the end of each scheduling time slot while the transmitted grids will be masked for the collaborators to avoid retransmission.

In this work, we consider that the feature fusion network $\Phi_{\text {fus}}$ can be represented in the form
\begin{equation}
\mathbf{F}_{f}^{t+1}(x,y)=\mathbf{W}_{e}^{t}(x,y)\mathbf{F}_{f}^{t}(x,y)+\sum_{j \in \mathcal{N}}\lambda_{j}^{t}\mathbf{W}_{j}^{t}(x,y)\mathbf{F}_{j}^{t+1}(x,y),
\end{equation}
where $\mathbf{W}_{e}^{t} \in \mathbb{R}^{H \times W}$ and $\mathbf{W}_{j}^{t} \in \mathbb{R}^{H \times W}$ represents the fusion weight for the BEV features $\mathbf{F}_{f}^{t}$ of the ego vehicle and the BEV features $\mathbf{F}_{j}$ of the CAVs. Generally, collaborators with higher spatial confidence values will have higher fusion weights. Note that the DiscoNet model \cite{li2021learning} used in this paper as well as some other common fusion models such as F-Cooper \cite{chen2019f} and Where2comm \cite{hu2022where2comm} meet this prerequisite. Building on this, we design a heuristic feature selection rule based on the spatial confidence map. Intuitively, a large $\boldsymbol{\tau}_j^t(x, y)$ with a small $\boldsymbol{\tau}_e^t(x, y)$ means that the ego vehicle is unable to detect the target vehicle on the grid $(x, y)$ while its $j$-th collaborator believes that there is a high probability of the existence of a vehicle on the grid $(x, y)$. The grids satisfying the above condition should be given higher transmission priority than other grids. Based on this, the feature selection mask of the $j$-th collaborator at $t$-th scheduling slot can be obtained by
\begin{equation}
\mathbf{M}_{j}^t = \text{TOPK}\left(\boldsymbol{\tau}_j^t(x, y)^2\left[1-\boldsymbol{\tau}_e^0(x, y)\right]|B_j^t\right), j \in \mathcal{N},
\end{equation}
where $\text{TOPK}\left(\cdot|B_j^t\right)$ stands for selecting $B_j^t$ grids with the highest values, and then the corresponding grid on the map $\mathbf{M}_{j}^t$ is set to 1. $\boldsymbol{\tau}_e^0$ is used instead of $\boldsymbol{\tau}_e^t$ since $\boldsymbol{\tau}_e^t, t>0$ is not available for collaborators. But $\boldsymbol{\tau}_e^t, t>0$ will be utilized in the scheduling algorithm design at the Ego side, which will be introduced in the next subsection. It is worth noting that $\boldsymbol{\tau}_j^t(x, y)^2$ instead of $\boldsymbol{\tau}_j^t(x, y)$ was used in the calculation, giving more consideration to $\boldsymbol{\tau}_j^t(x, y)$ compared to $\boldsymbol{\tau}_e^0(x, y)$.

\subsection{Reinforcement Learning Formulation of User Scheduling}\label{AA}
With the feature selection method in Section III-A, the optimization problem in (\ref{prob1}) reduces to the sequential user scheduling problem as
\begin{subequations}
    \begin{align}
        \mathop {\max }\limits_{\left\{\lambda_j^t\right\}_{j \in \mathcal{N}}^{t \in \mathcal{T}}}  \quad & \text{AP}\left(\hat{O}_{\text {pred}}^T, O_{\text {true}}\right) 
        \\
        \ \textrm{ s.t. } \qquad
        & \lambda_j^t \in\{0,1\}, \forall j \in \mathcal{N}, t \in \mathcal{T},  \\
        & \sum_{j \in \mathcal{N}} \lambda_j^t \leq 1, \forall t \in \mathcal{T}.
    \end{align}
\end{subequations}

\begin{figure}[t]
    \centering
    \subfigure[Spatial confidence map]{
        \includegraphics[width=0.232\textwidth]{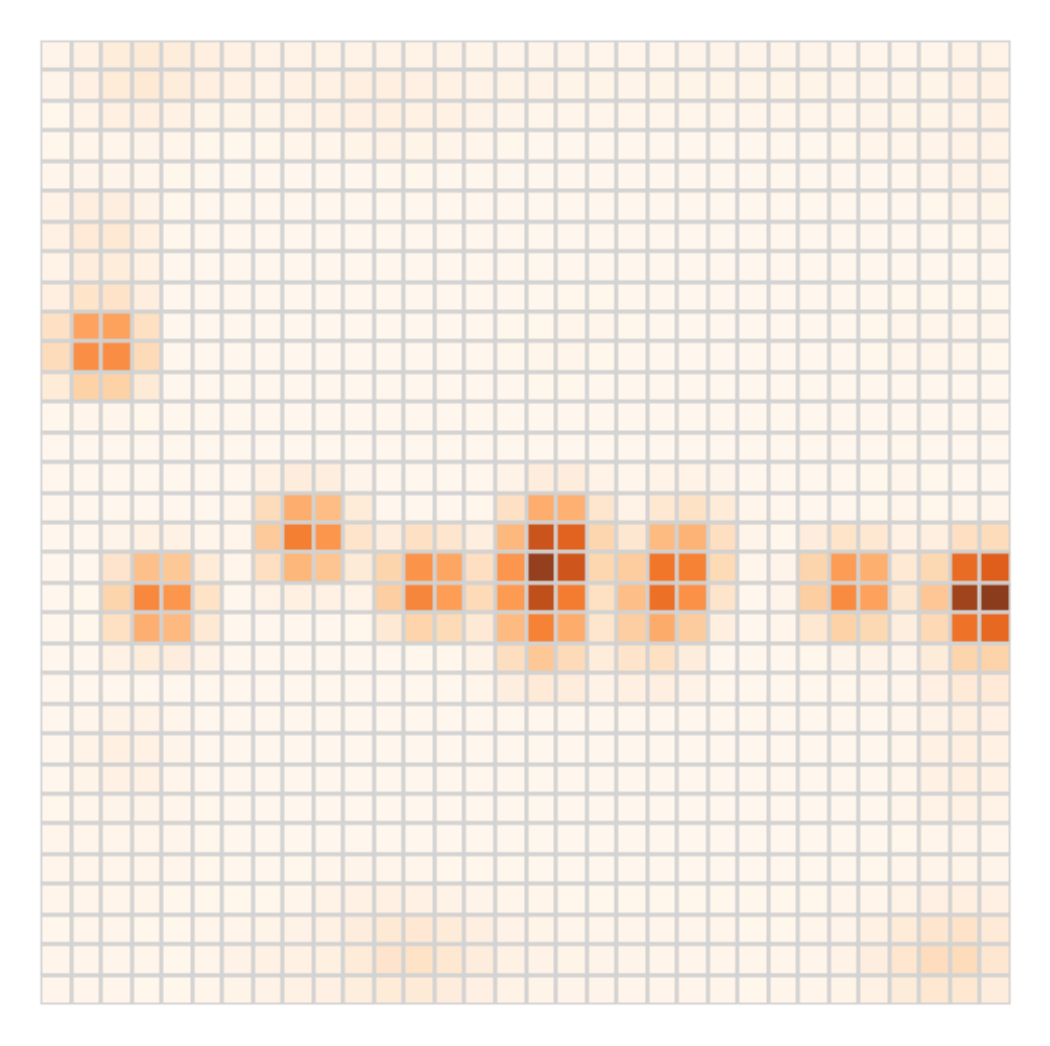}
    }
    \subfigure[Detection results]{
        \includegraphics[width=0.223\textwidth]{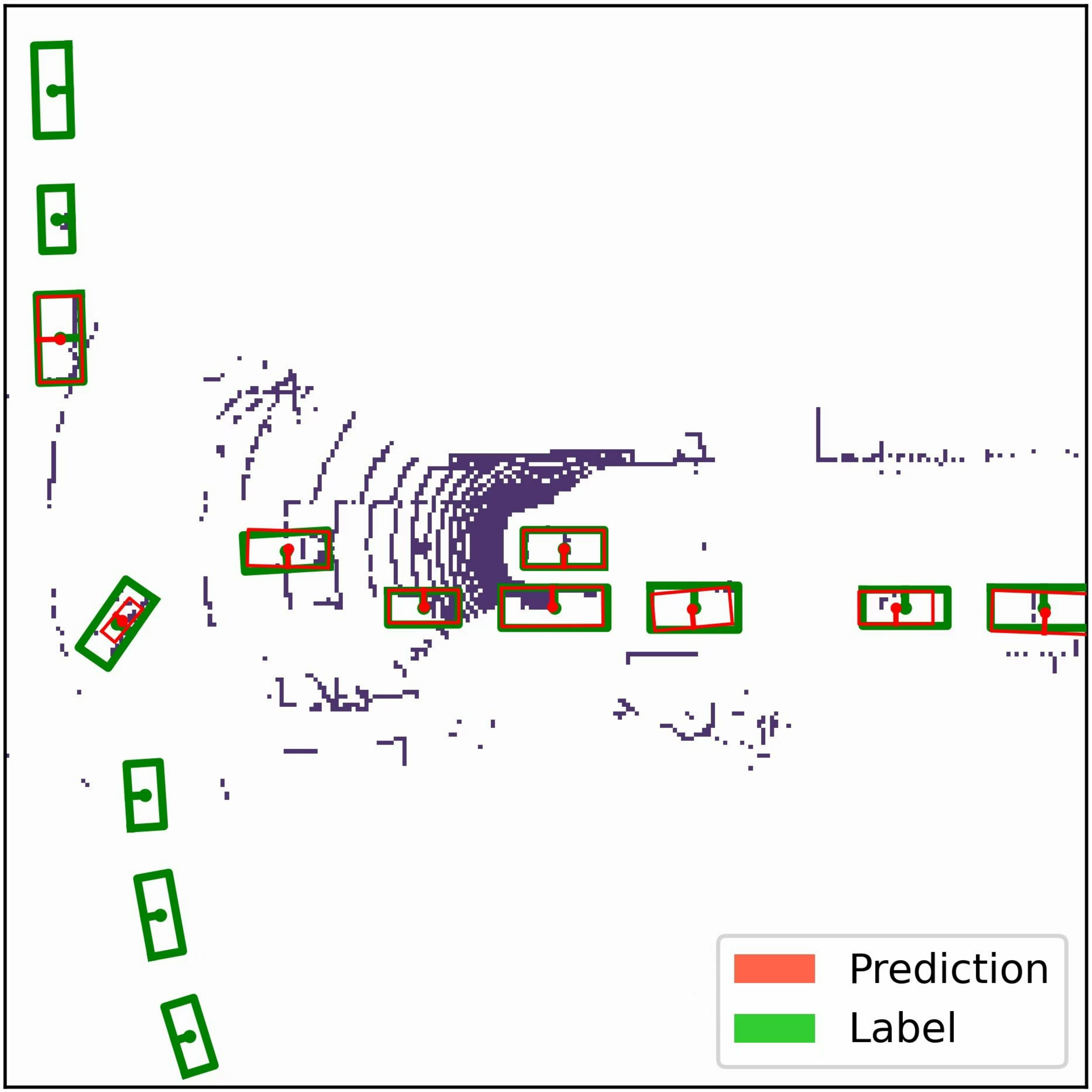}
    }
    \caption{Relationship between spatial confidence map and detection results.}
    \label{fig:twofigures}
\end{figure}

As it involves the fusion and extraction of BEV features, this optimization problem is difficult to solve by traditional optimization methods. Therefore, we propose in this work to solve this problem by resorting to DRL. First, the ego vehicle broadcasts the pose information to all collaborators, and then they align the features to the same coordinate system and transmit the spatial confidence map to the ego. Considering the huge dimensional gap between $\boldsymbol{\tau}_j^0 \in \mathbb{R}^{H \times W}$ and $\mathbf{F}_j \in \mathbb{R}^{H \times W \times C}$, the overhead for transmitting $\boldsymbol{\tau}_j^0, j \in \mathcal{N}$ is negligible. Based on the spatial confidence map and CSI, the ego vehicle makes sequential scheduling decisions $\lambda_j^t \in\{0,1\}, j \in \mathcal{N}$ at each slot to improve the perception performance of the ego vehicle. In this problem formulation based on reinforcement learning (RL), the ego vehicle acts as an agent and learns a good scheduling strategy through interaction with the environment \cite{schulman2017proximal}. We set the episode in RL as the sensor sampling interval $T$, and each step in an episode as the duration of a scheduling slot $t_0$. Specifically, the corresponding elements in RL are designed as follows: \\
\textbf{1) State space $\mathcal{S}$:} Intuitively, the choice of which link accesses the spectrum is influenced by two factors:
\begin{itemize}
    \item \textbf{The channel condition of the V2V link or V2I link:} When the channel condition is better, the ego vehicle can receive more features and achieve higher accuracy.
    \item \textbf{The importance of the perceptual information from different collaborators.} Features that hold higher priority are more essential for the ego vehicle’s decision-making process. In addition, when a collaborator has transmitted many features, its contribution to the ego vehicle perception gradually becomes smaller.
\end{itemize}
In light of these considerations, the state is designed as
\begin{equation}\label{state}
S_t=\left\{\operatorname{sum}\left(\mathbf{R}_{j}^t{ }^2\right), \max \left(\mathbf{R}_{j}^t{ }^2\right),\left\{\alpha_j, h_j^{t, 0}\right\}\right\}_{j \in \mathcal{N}},
\end{equation}
where $\mathbf{R}_{j}^t{ }$ is obtained by replacing $\boldsymbol{\tau}_e^0(x, y)$ in $\mathbf{M}_{j}^t$ by $\boldsymbol{\tau}_e^t(x, y)$ since $\boldsymbol{\tau}_e^t(x, y)$ is available at ego. ${ \mathbf{R}_{j}^t}^2$ stands for $\mathbf{R}_{j}^t \odot \mathbf{R}_{j}^t$. $\operatorname{sum}\left({\mathbf{R}_{j}^t}^2\right)$ means the summation of ${ \mathbf{R}_{j}^t}^2\left(x,y\right)$ for all $(x,y)$ and $\operatorname{max}\left({ \mathbf{R}_{j}^t}^2\right)$ represents the maximum value in ${ \mathbf{R}_{j}^t}^2$.

After the $t$-th scheduling slot, the agent experiences the state transitions of the environment in three aspects:
\begin{itemize}
    \item The update of the spatial confidence map from $j$-th collaborator at the receiver: $\boldsymbol{\tau}_j^{t+1}=\boldsymbol{\tau}_j^t \odot \left(1 - \mathbf{M}_{j}^{t}\right)$,
    \item The update of the spatial confidence map of the ego vehicle: $\boldsymbol{\tau}_e^{t+1}=\Phi_{\text {cls}}\left(\Phi_{\text {det}}\left(\mathbf{F}_f^{t}\right)\right)$
    \item The small-scale fading update follows the first-order Markov process: $h_j^{t+1,0}=\mu h_j^{t, T_s}+e, j \in \mathcal{N}$.
\end{itemize}
\textbf{2) Action space $\mathcal{A}$:} The action of the agent is whether each V2V link or V2I link accesses the spectrum or not, i.e., $A_t = {\{ \lambda_j^t \}}_{j\in\mathcal{N}}$, where $\lambda_j^t$ is a binary variable. \\
\textbf{3) Reward design:} We craft the reward to guide the agent towards optimal solutions aiming at maximizing the perception performance of the ego vehicle. According to the optimization objective (\ref{obj1}), the improvement of the AP perception metric after each scheduling slot can be incorporated into the reward. However, the improvement of the AP metric may be very sparse and its calculation is very time-consuming. Therefore, we use $\Delta \mathcal{L}_{\text {det}}^t=\mathcal{L}_{\text {det}}^{t-1}-\mathcal{L}_{\text {det}}^t$ to replace AP metric, where $\mathcal{L}_{\text {det}}$ is the loss for training the whole collaborative system, which will be expanded in the next section in detail. However, the diversity of perceptual scenarios may cause reward instability if $\Delta \mathcal{L}_{\text {det}}^t$ is directly designed as the reward, leading to very unstable training processes for the agent. To mitigate this issue and considering that a high transmission rate helps with the improvement of detection performance, we design the reward as the weighted sum of the transmission rate and the reduction of detection loss, i.e.,
\begin{equation}\label{reward}
\begin{aligned}
     R_{t} & = \lambda_{\text {r,l}} C_{\text {r}}^t+\lambda_{\text {det}} \Delta \mathcal{L}_{\text {det}}^t, \\
\end{aligned}
\end{equation}
where the weights $\lambda_{\text {r,l}}$ and $\lambda_{\text {det}}$ are introduced to balance the two components in the reward.

\subsection{Label-Free Design for User Scheduling}

Obviously, the reward in (\ref{reward}) requires detection labels. However, in real autonomous driving scenarios, it is difficult to obtain labels in real time for online reinforcement learning, and the annotation process of the bounding boxes consumes a lot of manpower and time. Therefore, this kind of reward design with labels can only be used in scenarios requiring sophisticated design and training of the model. Considering the non-analytical nature of the AP metric and challenges associated with obtaining label $O_{\text {true}}$, we make preliminary efforts to transform (\ref{obj1}) into a heuristic alternative that does not require labels, predicated on some characteristics of 3D object detection and devise the label-free reward according to the new objective.

In general, the loss function used to train the collaborative perception system $\mathcal{L}_{\text {det}}$ consists of three components, that is
\begin{equation}
\mathcal{L}_{\text {det}}=\lambda_{\text {cls}} \mathcal{L}_{\text {cls}}+\lambda_{\text {loc}} \mathcal{L}_{\text {loc}}+\lambda_{\text {dir}} \mathcal{L}_{\text {dir}},
\end{equation}
where $\mathcal{L}_{\text {loc}}$ and $\mathcal{L}_{\text {dir}}$ represent the localization loss and direction loss, respectively. These two parts are utilized to guide the module towards accurate position prediction and direction estimation. $\mathcal{L}_{\text {cls}}$ represents the classification loss, which quantifies the discrepancy between the confidence map $\boldsymbol{\tau}$ and the classification label $\mathbf{O}_{\text {cls}}$. For collaborative perception systems, the system with lower detection loss $\mathcal{L}_{\text {det}}$ tends to achieve higher perception accuracy, specifically in terms of the AP metric. Moreover, the collaboration between different units contributes to lower detection loss compared with stand-alone perception, mainly by solving the occlusion problem to decrease the classification loss $\mathcal{L}_{\text {cls}}$ during the training stage.

\newtheorem{assumption}{\textbf{Observation}}[section]
\begin{assumption}\label{as1}
Denote the detection results generated through the decoding and fusion network at the end of the $t$-th scheduling slot as $\hat{O}^t_{\text {pred}}$, i.e., $\hat{O}_{\text {pred}}^t=\Phi_{\text {pred}}\left(\Phi_{\text {dec}}\left(\mathbf{F}_f^t\right)\right)$. In a well-behaved collaborative perception system, the relationship among changes in performance metrics and losses can be expressed as follows: $\Delta \text{AP}^t \propto \Delta \mathcal{L}_{\text {det}}^t \propto \max \left(\Delta \mathcal{L}_{\text {cls}}^t, \varrho\right)$, where $\Delta \text{AP}^t=\text{AP}^t-\text{AP}^{t-1}, \Delta \mathcal{L}_{\text {det}}^t=\mathcal{L}_{\text {det}}^{t-1}-\mathcal{L}_{\text {det}}^t, \Delta \mathcal{L}_{\text {cls}}^t=\mathcal{L}_{\text {cls}}^{t-1}-\mathcal{L}_{\text {cls}}^t$ and $\varrho$ represents a very small positive number. Therefore, the following inequalities always hold: $\text{AP}^t \geq \text{AP}^{t-1}$ and $\mathcal{L}_{\text {det}}^t \leq \mathcal{L}_{\text {det}}^{t-1}$. In fact, this observation occurs in most practical scenarios.
\end{assumption}

From \textit{Observation \ref{as1}}, we have the following objective transformation: $\max \ \Delta \text{AP}^t \Rightarrow \max \ \Delta \mathcal{L}_{\text {det}}^t \Rightarrow \max \ \Delta \mathcal{L}_{\text {cls}}^t$, and the equivalence of maximizing $\Delta \text{AP}^t$ and $\Delta \mathcal{L}_{\text {det}}^t$ has been partially verified in \cite{chen2020ap}. The expression for $\mathcal{L}_{\text {cls}}$ is given by
\begin{equation}\label{loss_cls}
\mathcal{L}_{\text {cls}}=\sum_{(x, y)}-\boldsymbol{\alpha}(x, y)\left(1-\mathbf{p}(x, y)\right)^\beta \log \mathbf{p}(x, y),
\end{equation}
where $\beta$ is an additional hyperparameter to optimize the focus of the model, and the relationship between $p^a$, $\boldsymbol{\alpha}^a$ and $\boldsymbol{\tau}$ follows
\begin{equation}
\centering
\begin{aligned}\label{conf_p}
& \mathbf{p}(x, y)=\left\{\begin{array}{lc}
\boldsymbol{\tau}(x, y), & \text{if} \; \mathbf{O}_{\text {cls}}(x, y)=1, \\
1-\boldsymbol{\tau}(x, y), & \text{otheriwse,}
\end{array}\right. \\
& \boldsymbol{\alpha}(x, y)=\left\{\begin{array}{lc}
\eta, & \qquad\text{if} \; \mathbf{O}_{\text {cls}}(x, y)=1, \\
1-\eta, & \qquad \text{otheriwse,}
\end{array}\right. \\
\end{aligned}
\end{equation}
where $\mathbf{O}_{\text {cls}} \in \mathbb{R}^{H \times W}$ is the classification label in $O_{\text {true}}$ and $\eta$ is utilized for dealing with category imbalances.

According to (\ref{loss_cls}) and (\ref{conf_p}), it is expected that the gap between confidence value $\boldsymbol{\boldsymbol{\tau}}(x,y)$ and $\mathbf{O}_{\text {cls}}(x, y)$ is small, leading to low classification loss $\mathcal{L}_{\text {cls}}$ during the training phase. For the implementation phase, there exists a threshold $\zeta$ for classification prediction, that is, anchors with confidence $\boldsymbol{\boldsymbol{\tau}}(x, y) > \zeta$ are considered positive, i.e., $\mathbf{\hat{O}}_{\text {cls}}(x, y)=1$ and otherwise considered negative.

\begin{assumption}\label{as2}
If $\left(\boldsymbol{\boldsymbol{\tau}}_e^t(x, y)-\zeta\right)\left(\mathbf{O}_{\text {cls}}(x, y)-\zeta\right)>0, \exists t \in \mathcal{T}$ , then $(\boldsymbol{\boldsymbol{\tau}}_e^{t^{\prime}}(x, y)-\zeta)\left(\mathbf{O}_{\text {cls}}(x, y)-\zeta\right)>0, \forall t^{\prime} \geq t$.
\end{assumption}

To verify this observation, we conduct an experimental study on the V2X-Sim dataset \cite{li2022v2x}. As shown in Table I, the probability of True-to-False cases is less than 1\% even with only two collaborators, and the probability becomes lower as the number of collaborators increases. From \textit{Observation \ref{as2}}, we can derive that collaboration between collaborators and the ego vehicle is very unlikely to lead to a True-to-False change in the ego vehicle's prediction. Thus prediction always undergoes False-to-True shifts, thereby contributing to the reduction of classification loss $\mathcal{L}_{\text {cls}}$. Based on \textit{Observation \ref{as1}} and \textit{Observation \ref{as2}}, the original objective (\ref{obj1}) can be transformed into
\begin{equation}\label{obj3}
\begin{aligned}
    \text{AP} & \left( \hat{O}_{\text {pred}}^T, O_{\text {true}}\right) \Rightarrow \sum_{t=1}^T \Delta \mathcal{L}_{\text {det}}^t \Rightarrow \sum_{t=1}^T \Delta \mathcal{L}_{\text {cls}}^t\\
    & \Rightarrow \sum_{t=1}^T \sum_{(x, y)} \mathbb{I}\left[\left(\boldsymbol{\boldsymbol{\tau}}_e^t(x, y)-\zeta\right) \left(\boldsymbol{\boldsymbol{\tau}}_e^{t-1}(x, y)-\zeta\right) < 0\right],
\end{aligned}
\end{equation}
where $\mathbb{I}\left[\left(\boldsymbol{\boldsymbol{\tau}}_e^t(x, y)-\zeta\right) \left(\boldsymbol{\boldsymbol{\tau}}_e^{t-1}(x, y)-\zeta\right) < 0\right]$ represents whether the prediction changes for two adjacent slots. Although the new objective in (\ref{obj3}) can be calculated without labels, it will be too sparse due to the sparsity of prediction change. One solution is to use the measure of the gap between $\boldsymbol{\boldsymbol{\tau}}_e^{t}(x, y)$ and $\boldsymbol{\boldsymbol{\tau}}_e^{t-1}(x, y)$, such as $\left| \boldsymbol{\boldsymbol{\tau}}_e^t(x, y)-\boldsymbol{\boldsymbol{\tau}}_e^{t-1}(x, y)\right|^2$, as the heuristic objective. Nevertheless, the larger gap between $\boldsymbol{\boldsymbol{\tau}}_e^{t}(x, y)$ and $\boldsymbol{\boldsymbol{\tau}}_e^{t-1}(x, y)$ does not necessarily contribute to lower classification loss and further detection loss.

\begin{assumption}\label{as3}
For well-behaved collaborative perception, when $\left(\boldsymbol{\boldsymbol{\tau}}_e^t(x, y)-\boldsymbol{\boldsymbol{\tau}}_e^{t-1}(x, y)\right)\left(\mathbf{O}_{\text {cls}}(x, y)-\boldsymbol{\boldsymbol{\tau}}_e^{t-1}(x, y)\right)<0$ occurs, $\left|\boldsymbol{\boldsymbol{\tau}}_e^t(x, y)-\boldsymbol{\boldsymbol{\tau}}_e^{t-1}(x, y)\right|<\sqrt{\xi}$ always satisfies. where $\xi$ is a small positive threshold.
\end{assumption}

% \begin{figure*}[htbp]
% \centering
% \includegraphics[width=0.88\linewidth]{Scheduling_Model_2.pdf}
% \caption{An illustrative architecture of user scheduling for V2X-aided collaborative perception based on spatial confidence map.}
% \end{figure*}

The meaning of this observation is: for a well-behaved system, as $\boldsymbol{\boldsymbol{\tau}}_e^t(x, y)$ grows further away from $\mathbf{O}_{\text {cls}}(x, y)$, the change, i.e., $\left|\boldsymbol{\boldsymbol{\tau}}_e^t(x, y)-\boldsymbol{\boldsymbol{\tau}}_e^{t-1}(x, y)\right|$ is less than the threshold $\sqrt{\xi}$, demonstrating the robustness of the system. Similar to \textit{Observation \ref{as2}}, the \textit{Observation \ref{as3}} is also verified experimentally as shown in Table I. According to Table I, the value of $\zeta$ in this paper is set as 0.05 such that the corresponding probability of \textit{Observation \ref{as3}} not happening is less than 1\%. When $\boldsymbol{\boldsymbol{\tau}}_e^t(x, y)$ is relatively large and $\boldsymbol{\boldsymbol{\tau}}_j^t(x, y)$ is relatively small, there may be a slight decrease compared with $\boldsymbol{\boldsymbol{\tau}}_e^{t+1}(x, y)$. As a result, classification performance will rise slightly. However, according to $\Delta \text{AP}^t \propto \Delta \mathcal{L}_{\text {det}}^t \propto \max \left(\Delta \mathcal{L}_{\text {cls}}^t, \varrho\right)$, when $\Delta \mathcal{L}_{\text {cls}}^t \leq 0$, its impact on the perception performance is negligible. Combining the \textit{Observation \ref{as3}} and objective in (\ref{obj3}), the value of $\left|\boldsymbol{\boldsymbol{\tau}}_e^t(x, y)-\boldsymbol{\boldsymbol{\tau}}_e^{t-1}(x, y)\right|$ should be maximized in order to induce a change in prediction. The optimization problem can be further transformed into
\begin{subequations} \label{final_ques}
    \begin{align}
        \mathop {\max }\limits_{\left\{\lambda_j^t\right\}_{j \in \mathcal{N}}^{t \in \mathcal{T}}}  \quad & \sum_{t=1}^T \sum_{(x, y)} \max \left[T^t(x, y), G^t(x, y)\right] \label{obj2}
        \\
        \ \textrm{ s.t. } \qquad
        & \lambda_j^t \in\{0,1\}, \forall j \in \mathcal{N}, t \in \mathcal{T},  \\
        & \sum_{j \in \mathcal{N}} \lambda_j^t \leq 1, \forall t \in \mathcal{T},
    \end{align}
\end{subequations}
where $T^t(x, y)=\mathbb{I}\left[\left(\boldsymbol{\boldsymbol{\tau}}_e^t(x, y)-\zeta\right)\left(\boldsymbol{\boldsymbol{\tau}}_e^{t-1}(x, y)-\zeta\right)<0\right]$, $G^t(x, y)=g\left(\left| \boldsymbol{\boldsymbol{\tau}}_e^t(x, y)-\boldsymbol{\boldsymbol{\tau}}_e^{t-1}(x, y)\right|^2 - \xi\right)$ with $g$ being the ReLU activation function \cite{glorot2011deep}. The essence of this optimization goal is to maximize the absolute difference $\left| \boldsymbol{\boldsymbol{\tau}}_e^t(x, y)-\boldsymbol{\boldsymbol{\tau}}_e^{t-1}(x, y)\right|$. When this difference is substantial enough for a beneficial prediction shift to occur, then $T^t(x, y) = 1$, thereby maximizing the objective value. This heuristic objective encourages changes in confidence values, thereby increasing correct predictions. The consistency of (\ref{obj1}) and (\ref{obj2}) will be verified in Section IV.

In this formulation, $T^t(x, y)$ and $G^t(x, y)$ are both related to the spatial confidence map $\boldsymbol{\tau}_i^t$, for which there is no explicit expression since it is generated by the neural network $\Phi_{\text {gen}}$. Consequently, traditional optimization methods encounter significant challenges in this scenario \cite{liang2019deep}. Recall in Problem (\ref{final_ques}), the original optimization objective is replaced by a summation of the utility value, i.e., $\sum_{(x, y)} \max \left[T^t(x, y), G^t(x, y)\right]$, in each scheduling slot, which makes the problem suitable for use with RL. Considering that the AP metric has been approximately transformed into an optimization objective without labels and that each element involved in the new optimization objective is available in real time on the ego side, the label-free reward is designed as
\begin{equation}\label{reward2}
R_{t}=\lambda_{\text {r,nl}} C_{\text {r}}^t+\lambda_{\text {u}} \sum_{(x, y) \in \Xi^t} \max \left[T^t(x, y), G^t(x, y)\right],
\end{equation}
where $\Xi^t=\left\{(x, y) \mid \mathbf{M}_{j}^t(x, y)=1, \lambda_j^t=1\right\}$ represents the set of all grids transmitted in the $t$-th scheduling slot.

\subsection{Proposed DDQN-Based Scheduling Algorithm}

\begin{table}[t]
  \begin{center}
    \caption{True-to-False Probability and Violation Probability}
    \setlength{\tabcolsep}{6pt}
    \vspace{-0.6cm}
    \scalebox{1.2}{
        \begin{tabular}[t]{c|c|c|c|c}
            \hline
            \textbf{Collaborator Number} & 2 & 3 & 4 & 5 \\
            \hline
            \textbf{True-to-False Prob (\%)} & 0.80 & 0.78 & 0.78 & 0.71 \\
            \hline
            \hline
            \textbf{Threshold $\zeta$} & 0.001 & 0.005 & 0.01 & 0.05 \\
            \hline
            \textbf{Violation Prob (\%)} & 6.9 & 1.9 & 1.3 & 0.5 \\
            \hline
        \end{tabular}
    }
  \end{center}
\end{table}

\begin{figure}[t]
\centering
% \vspace{-0.4cm}
\includegraphics[width=\linewidth]{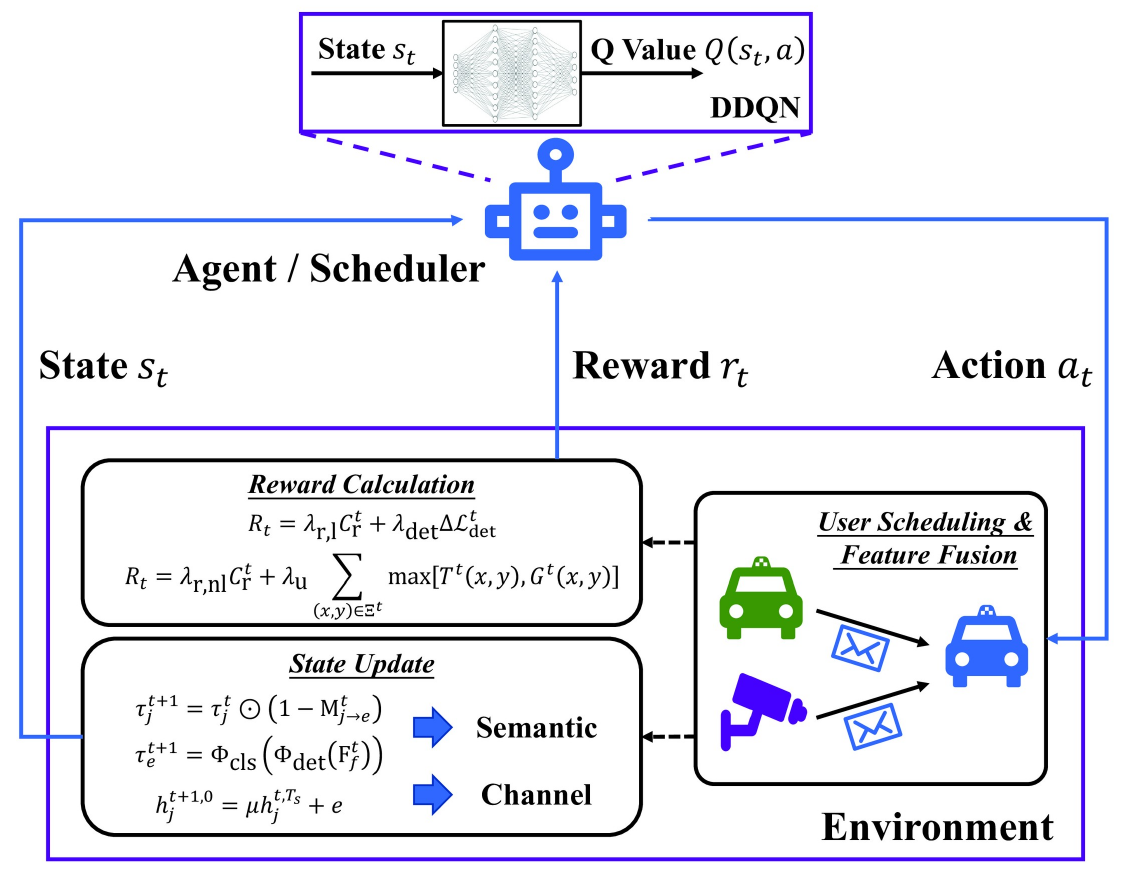}
\caption{Illustration of the scheduler as an agent interacting with the environment.}
\end{figure}

In this work, we leverage the DDQN algorithm \cite{van2016deep} with memory replay for training the agent for adaptive sequential scheduling decisions on the ego side by interacting with the environment as depicted in Fig. 5. Specifically, the goal of training the agent is to select a series of actions to maximize the return on average, denoted as $G_t$, i.e., the cumulative discounted reward as
\begin{equation}
G_t=\sum_{k=0}^\infty\gamma^kR_{t+k+1},\quad0\leq\gamma\leq1,
\end{equation}
where $\gamma$ represents the discount factor. In Q-learning, the Q function with state $a$ taken under state $s$ for a given policy can be defined as
\begin{equation}
Q(s, a)=\mathbb{E}\left[G_t \mid s_t=s, a_t=a\right].
\end{equation}
For the DDQN algorithm, the Q-values $Q(s, a)$ for all state-action pair $(s, a)$ are estimated by a Q-network with the parameter $\theta$, denoted as $Q(s, a;\theta)$. Thus the essence of training the agent is to iteratively update the parameters of the Q-network by the collected experience so that the estimation of the Q-function value becomes more accurate, thus guiding the agent to make better decisions.

\begin{algorithm}[t]
\begin{minipage}{\textwidth}
    \SetAlgoLined %显示end
	\caption{SchedCP User Scheduling Framework for\\
    Collaborative Perception: Training Stage}%算法名字
	Initialize the paramater $\theta, \theta^{-}$.\\
    Pre-train the Pointpillars for 3D object detection.\\
	\For{each episode}{
		The LiDAR sensors acquire point cloud: $\chi_j, j \in \mathcal{N}$.\\
        All units generate the spatial confidence maps which \\ 
        are gathered by the ego: $\boldsymbol{\tau}_j^0=\Phi_{\text {gen}}\left(\mathbf{F}_j\right), j \in \mathcal{M}$.\\
        Update vehicle locations and large-scale fading $\alpha_i$. \\
        \For{each scheduling slot $t$}{
    		Obtain the observation $s_t$ according to (\ref{state}).\\
            Choose the action $a_t = \mathop{\arg\max}_{a} Q(s_t,a;\theta)$ \\
            with probability $1-\epsilon$ and otherwise choose \\
            an action $a_t$ randomly.\\
            \For{each sub-time slot $t_s$}{
                Update channel small-scale fading in (\ref{fast_fading}). \\
                Calculate the transmission rate $C_j^{t, t_s}$ and select \\
                features $\left\{\mathbf{F}_{j}^t\right\}_{\lambda_j^t=1, j \in \mathcal{N}}$ to transmit. 
            }
            Calculate the reward $r_t$ according to (\ref{reward}) or (\ref{reward2}). \\
            Obtain the new observation $s_{t+1}$. \\
            Store $(s_t,a_t,r_t,s_{t+1})$ in the replay memory.
    	}
        Sample a mini-batch of experiences $D$ from the replay \\
        buffer and update $\theta$ by optimizing error in (\ref{update}). \\
        Every $E$ steps reset $\theta^{-} \leftarrow \theta$.
	}
\end{minipage}
\end{algorithm}

Our DRL-based scheduling algorithm can be divided into two phases: training and implementation. The whole training procedure is presented in Algorithm 1. During training, the agent obtains an observation $s_t$ from the environment at each step $t$. It chooses an action $a_t$ from the predefined action space with $\epsilon$-greedy policy, meaning that with probability $1-\epsilon$ the action with maximal Q value is chosen, represented as
\begin{equation}
a_t=\arg \max _{a \in A} Q\left(s_t, a\right),
\end{equation}
while a random action is chosen with probability $\epsilon$. Following this action selection, the agent receives a reward $r_t$ given by the environment. Consequently, the environment state changes due to the action of the agent $a_t$ and the agent obtains a new observation $s_{t+1}$. In this way, the agent collects and stores the transition tuple $(s_t,a_t,r_t,s_{t+1})$ through interacting with the environment with some soft policies and uses the samples to update the deep Q-network (DQN). At the end of each episode, a mini-batch of experiences $\mathcal{D}$ are uniformly sampled from the replay memory for updating $\theta$ \cite{mnih2015human}, the parameter of the deep Q-network, by minimizing the loss $L$ as
\begin{equation}\label{update}
L = \sum_{\left(s_t, a_t\right) \in \mathcal{D}}\left[y_t-Q\left(s_t, a_t ; \theta\right)\right]^2,
\end{equation}
where $y_t$ is given by
\begin{equation}
y_t = r_{t}+\gamma \max _{a^{\prime}} Q\left(s_{t+1}, a^{\prime} ; \theta^{-}\right),
\end{equation}
with $\theta^-$ being the parameter of the target Q-network \cite{van2016deep}.

\begin{table}[t]
  \begin{center}
    \caption{Simulation Parameter}
    \setlength{\tabcolsep}{6pt}
    \vspace{-0.8cm}
    \scalebox{1.3}{
        \begin{tabular}[t]{l|c}
            \hline
            \textbf{Parameter} & \textbf{Value} \\
            \hline
            Number of collaborators & 4 \\
            Absolute vehicle speed & 0 km/h $\sim$ 25 km/h \\
            Carrier frequency & 5.9 GHz \\
            Sensor sampling interval $T$ & 200 ms \\
            Large-scale fading update & 200 ms \\
            Small-scale fading update & 1 ms \\
            Length of scheduling slot & 5 ms \\
            Transmit power & 23 dBm \\
            Bandwidth & 200 kHz $\sim$ 600 kHz \\
            Vehicle antenna gain & 3 dBi \\
            Vehicle receiver noise figure & 9 dB \\
            Noise power spectral density & -174 dbm/Hz \\
            Path loss model & LOS in WINNER +  \\
            & B1 Manhattan \\
            Shadowing distribution & Log-normal \\
            Shadowing standard deviation & 3 dB \\
            Decorrelation distance & 10 m \\ 
            \hline
        \end{tabular}
    }
  \end{center}
\end{table}
\begin{table}[t]
  \begin{center}
    \caption{Parameter for Model Training}
    \setlength{\tabcolsep}{6pt}
    \vspace{-0.8cm}
    \scalebox{1.24}{
        \begin{tabular}[t]{l|c}
            \hline
            \textbf{Parameter} & \textbf{Value} \\
            \hline
            Training frames & 2574 \\
            Training episodes & (20000 / 30000) \\
            Testing episodes & 500 \\
            $\left(\lambda_{\text {r,l}}, \lambda_{\text {det}}\right)$ & (0.02, 8)  \\
            $\left(\lambda_{\text {r,nl}}, \lambda_{\text {u}}\right)$ & (0.04, 0.3) \\
            Architecture of DDQN & Three-layer fully connected \\
            & neural network with 500, \\
            & 250, 125 neurons, respectively \\
            \hline
        \end{tabular}
    }
  \end{center}
\end{table}

Notice that the state transition is related to the feature selection rule $\Phi_{\text {sel}}$. Thus SchedCP can actually adapt the V2X user scheduling decisions according to different feature selection rules in real time through online RL, which is difficult to accomplish with traditional optimization methods.

\section{Simulation Results}

In this section, we first verify some of the previous hypotheses through simulation experiments. We then conduct experiments to verify the effectiveness and robustness of our proposed SchedCP framework. Finally, we illustrate the working mechanism of the proposed algorithm through a case study. All experiments are conducted on the V2X-Sim dataset, which is the first synthetic V2X-aided collaborative perception dataset in autonomous driving generated using CARLA \cite{dosovitskiy2017carla} and SUMO \cite{lopez2018microscopic}. The collaborative perception backbone we use is DiscoNet \cite{li2021learning}. Considering that the DiscoNet model does not consider non-ideal communication environments, we utilize the sensor data from the V2X-Sim dataset and build a communication simulator according to 3GPP TR 38.885 \cite{3rd2019technical} to simulate the communication environment. The main simulation parameters are presented in Table II.

As listed in Table III, we use 2,574 frames from the dataset in the training phase and 500 in the testing phase. The numbers of training episodes are set to 20,000 for the label-dependent reward and 30,000 for the label-free reward. We obtain our proposed models using the rewards in (\ref{reward}) and (\ref{reward2}), denoted as SchedCP-wl and SchedCP, respectively. The exploration probability $\epsilon$ of the $\epsilon$-greedy strategy decreases linearly from 1 to 0.02 for the first 16,000 episodes and remains unchanged thereafter. In addition, the parameters of the Q network are copied to the target network every ten episodes.

\subsection{Convergence Behavior}

\begin{figure}[t]
\centering
% \vspace{-0.4cm}
\includegraphics[scale=0.59]{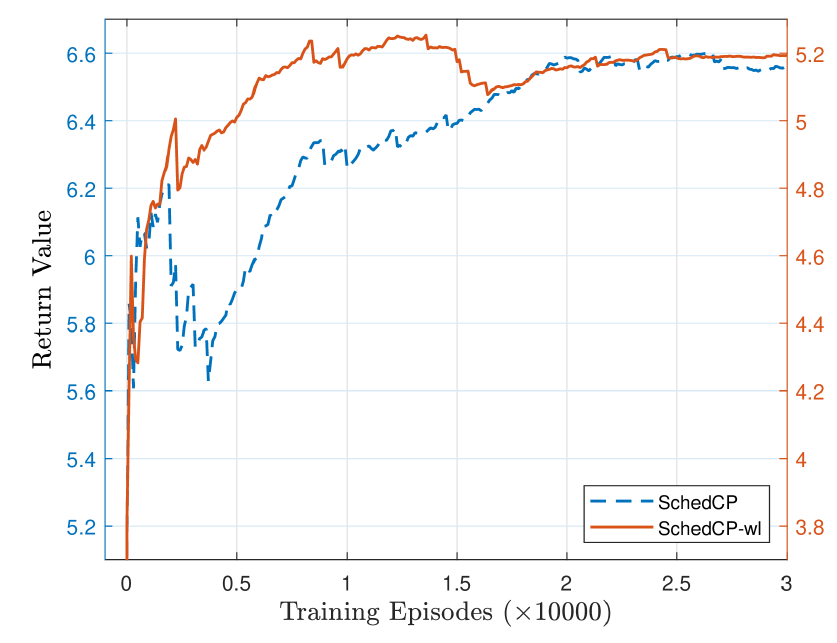}
\caption{Return on prepared validation set with increasing iterations.}
\end{figure}

Fig. 6 shows the return curves of both label-dependent and label-free scenarios. The left y-axis corresponds to the return value of SchedCP-wl, and the right y-axis corresponds to the return value of SchedCP. Considering the diversity of autonomous driving scenarios, which leads to the variability of returns in different episodes in RL, we adopt the method of running the model on 15 randomly selected frames from a small validation dataset every 100 episodes to compute the returns and thus plot the return curves. As Fig. 6 demonstrates, the model converges within approximately 10,000 episodes with the reward according to (\ref{reward}), while the model requires about 20,000 episodes with the reward according to (\ref{reward2}). This discrepancy can be attributed to the more precise guidance that labels offer during model training compared to the heuristic utility values.

\subsection{Perception Performance Comparison}

\begin{figure} 
    \centering 
    \subfigure[AP@0.50]{\label{fig:subfig:a}
    \includegraphics[scale=0.57]{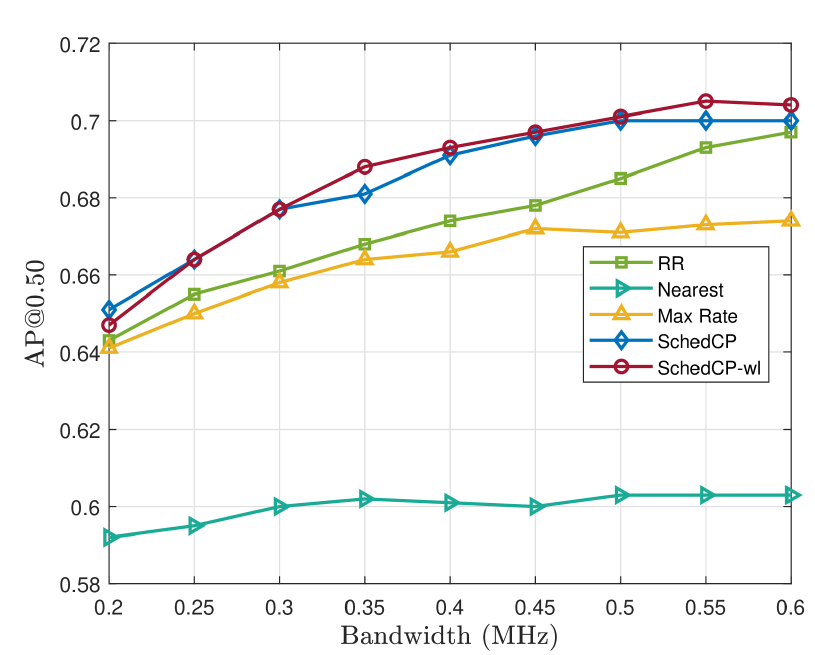}}
    \vfill
    \subfigure[AP@0.70]{\label{fig:subfig:a}
    \includegraphics[scale=0.57]{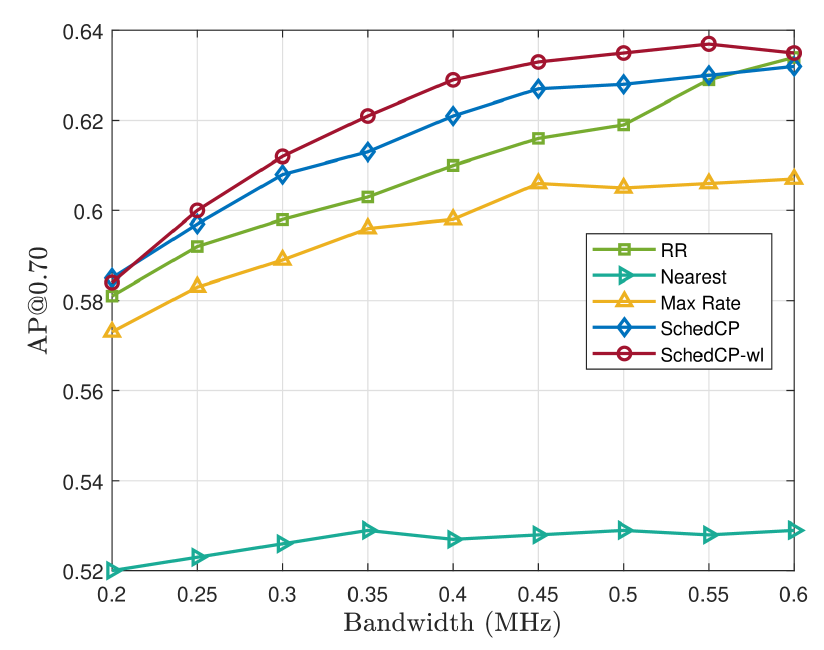}}
    \caption{Performance under different bandwidth conditions in terms of AP@0.50 and AP@0.70.}
    \label{fig:subfig}
\end{figure}

In this experiment, we compare the perception performance of different scheduling strategies under different bandwidths. Our models are trained under the bandwidth of 300 kHz and in the testing phase the bandwidth varies from 200 kHz to 600 kHz. We compare the proposed algorithm with the following three baseline methods:
\begin{enumerate}
    \item \textbf{Nearest scheduling (Nearest):} The link to the nearest CAV accesses the spectrum.
    \item \textbf{Round robin scheduling (RR):} Different links take turns to access the spectrum.
    \item \textbf{Maximum rate scheduling (Max Rate):} The link with the maximum transmission rate at each time step accesses the spectrum.
\end{enumerate}

We first compare our methods with baseline schemes in terms of perception performance. As Fig. 7 illustrates, SchedCP-wl and SchedCP generally outperform baselines in terms of both AP@0.50 and AP@0.70. The AP of SchedCP is only a little lower than that of SchedCP-wl, demonstrating the effectiveness of the label-free reward design in (\ref{reward2}). From the comparison between Fig. 7(a) and Fig. 7(b), it can be observed that the performance gap between SchedCP and SchedCP-wl is larger with a larger IoU threshold. This is because a larger IoU threshold implies a higher requirement for the localization accuracy. The reward in (\ref{reward2}) is designed only from the perspective of classification accuracy, whereas the $\Delta \mathcal{L}_{\text {det}}^t$ part in the reward in (\ref{reward}) includes the localization loss $\Delta \mathcal{L}_{\text {loc}}^t$ thus enabling more accurate localization.

\begin{figure} 
    \centering 
    \subfigure[Detection loss]{\label{fig:subfig:a}
    \includegraphics[scale=0.57]{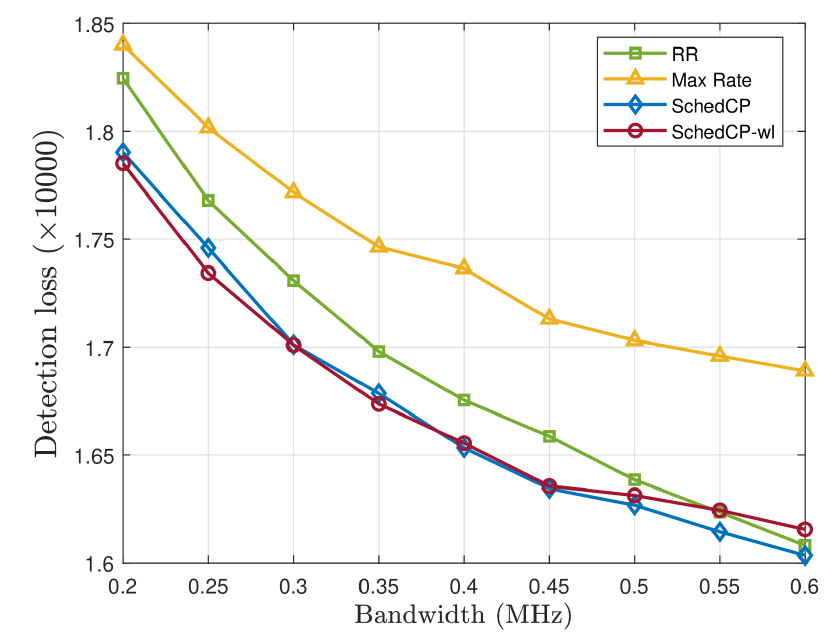}}
    \vfill
    \subfigure[Classification loss]{\label{fig:subfig:a}
    \includegraphics[scale=0.57]{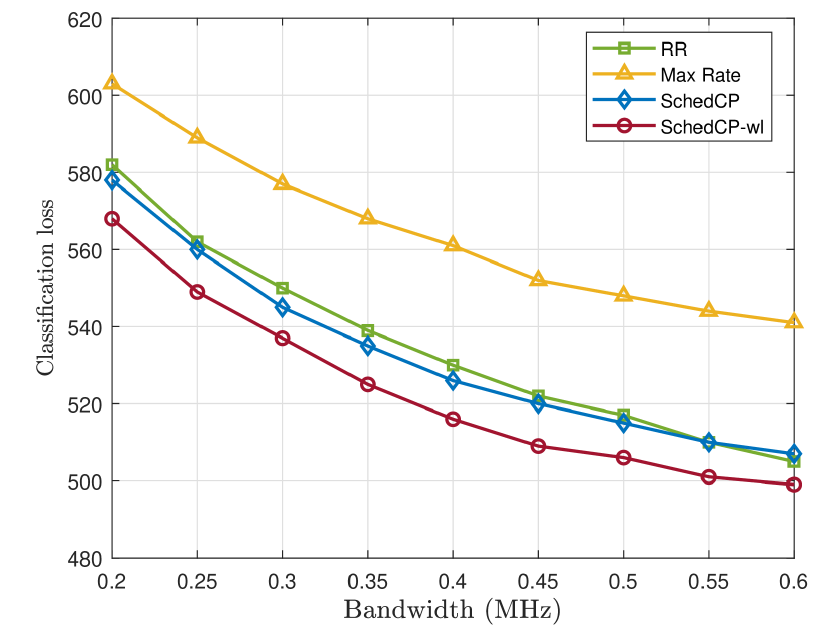}}
    \caption{Detection and classification loss under different bandwidth conditions.}
    \label{fig:subfig}
\end{figure}

As for the baseline methods, the performance of the Nearest scheduling method is much lower than the other methods in terms of the AP metric. Compared with the Nearest method, SchedCP improves by 12.9\% and 16.3\% in terms of AP@0.50 and AP@0.70 at 300 kHz, respectively. Since the Nearest method only schedules the link to the nearest collaborator each time, the ego vehicle can only acquire limited semantic information from the nearest CAV, which illustrates the importance of selecting links by taking different perspectives into account. Apart from the Nearest method, the Max Rate method also fails to work in the context of user scheduling in collaborative perception. Specifically, SchedCP-wl improves the perception performance metrics by 2.9\% and 4.0\% compared with the Max Rate method in terms of AP@0.50 and AP@0.70 at 300 kHz, which is a large improvement for autonomous driving. Furthermore for the Max Rate method, higher transmission rate can only result in limited perceptual gain since it is not intended for perceptual accuracy. Therefore the perceptual performance gap between the Max Rate and SchedCP method becomes even larger in high bandwidth regimes.

\begin{figure}[t]
\centering
\includegraphics[scale=0.57]{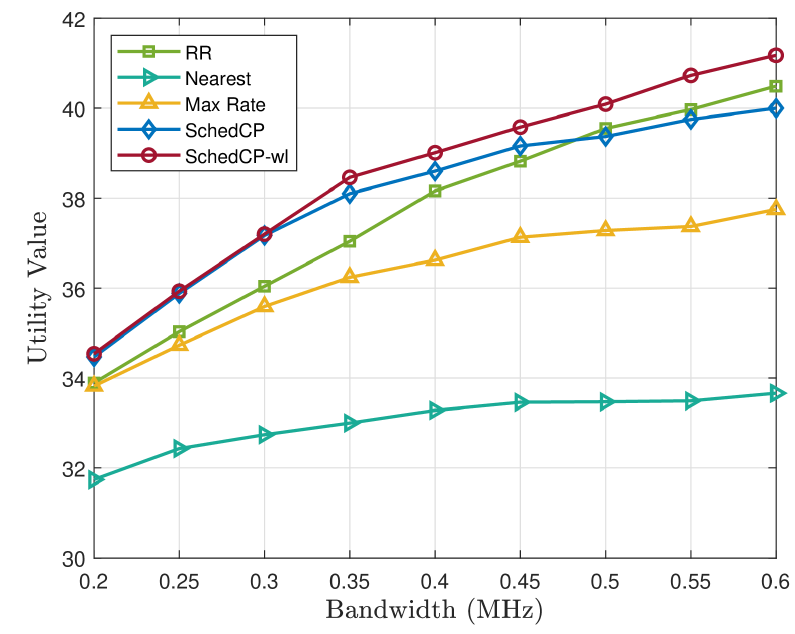}
\caption{ The utility values of our methods compared with other baselines under different bandwidth conditions.}
\end{figure}

Fig. 8 and Fig. 9 demonstrate the detection and classification loss as well as the utility values, i.e., the objective (\ref{obj2}) of the ego vehicle across different scheduling methods at different bandwidths. Comparing Fig. 7 and Fig. 8, we observe that the trend of detection loss $\mathcal{L}_{\text {det}}$ is closer to the opposite of the trend of AP performance compared with classification loss $\mathcal{L}_{\text {cls}}$ because detection loss is a more comprehensive measurement of perceptual accuracy, consisting of not only $\mathcal{L}_{\text {cls}}$ but also $\mathcal{L}_{\text {loc}}$ and $\mathcal{L}_{\text {dir}}$. As shown in Fig. 9, the trend of the utility values is generally similar to that of the AP in Fig. 7 except for the RR method. It can be found that from 500 kHz to 600 kHz, the RR method has higher utility value than the SchedCP-wl method while its AP performance is no better than that of the latter. This is because collaboration between different units at low bandwidth mainly solves the occlusion problem and improves the confidence values to reduce the classification loss $\mathcal{L}_{\text {cls}}$, while in high bandwidth collaboration mainly reduces $\mathcal{L}_{\text {loc}}$ and $\mathcal{L}_{\text {dir}}$ by improving the localization accuracy of the objects. Therefore, this heuristic utility function may deviate a little from the measurement of perceptual accuracy in high bandwidth regimes since it is derived from $\mathcal{L}_{\text {cls}}$, thus ignoring the effect of collaboration on the other two losses. Overall, although the heuristic utility value cannot completely replace the original AP metrics in evaluating perceptual accuracy, its corresponding reward in (\ref{reward2}) has guided the DDQN training process to a great extent.

\begin{table}[t]
\setlength{\tabcolsep}{4pt}
\centering
\caption{Transmission rate under different bandwidth conditions.}
\label{table1}
\begin{tabular}{|c|c|c|c|c|c|c|c|c|}
\hline
\multicolumn{4}{|c|}{\multirow{2}{*}{\textbf{Methods}}} & \multicolumn{5}{c|}{\textbf{Channel capacity (Mbps)}}\\
\cline{5-9}
\multicolumn{4}{|c|}{} & 200 kHz & 300 kHz & 400 kHz & 500 kHz & 600 kHz \\
\hline
\multicolumn{4}{|c|}{Nearest} & 3.82 & 5.56 & 7.25 & 8.92 & 10.54 \\
\hline
\multicolumn{4}{|c|}{RR} & 3.31 & 4.80 & 6.24 & 7.65 & 9.03 \\
\hline
\multicolumn{4}{|c|}{Max Rate} & 4.05 & 5.90 & 7.72 & 9.50 & 11.23 \\
\hline
\multicolumn{4}{|c|}{SchedCP} & 3.75 & 5.45 & 7.13 & 8.77 & 10.38 \\
\hline
\multicolumn{4}{|c|}{SchedCP-wl} & 3.62 & 5.28 & 6.89 & 8.48 & 10.03 \\
\hline
\end{tabular}
\label{table_MAP}
\end{table}

\begin{figure} 
    \centering 
    \subfigure[AP@0.50]{\label{fig:subfig:a}
    \hspace{-0.4cm} \includegraphics[scale=0.45]{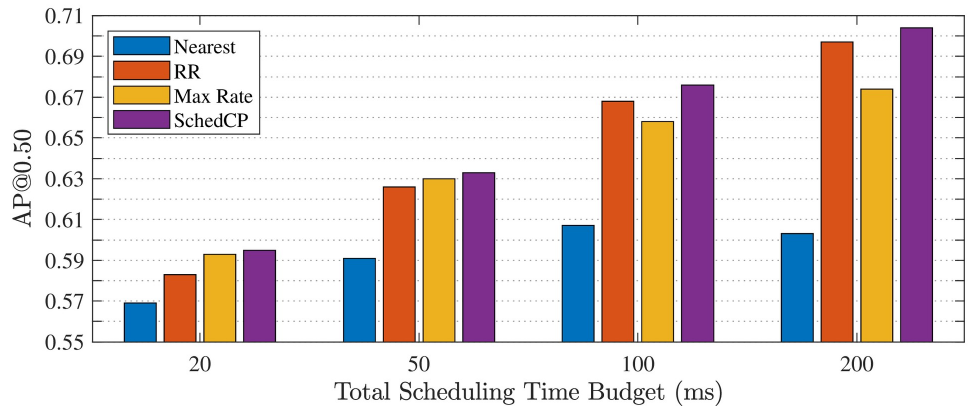}}
    \subfigure[AP@0.70]{\label{fig:subfig:a}
    \hspace{-0.4cm} \includegraphics[scale=0.45]{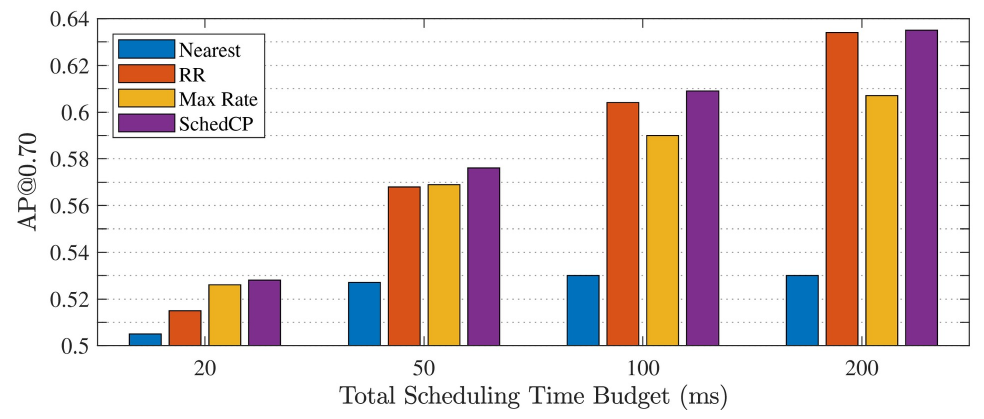}}
    \caption{Perception performance under different sensor sampling intervals in terms of AP@0.50 and AP@0.70.}
    \label{fig:subfig}
\end{figure}

Table IV illustrates the transmission rates under different scheduling strategies. It is observed that the transmission rate achieved by the RR method is significantly lower than that of other methods, which is a primary constraint of its performance. The transmission rate of the Nearest method is the second highest, which can be possibly explained by the fact that the path loss from the closest collaborator to the ego tends to be the smallest among all collaborators. Therefore the transmission rate of the corresponding link is relatively large. Although our methods are inferior to the Nearest and Max Rate methods in terms of transmission rates, they outperform these two methods in terms of AP metric. This demonstrates that SchedCP and SchedCP-wl develop an effective scheduling strategy, to achieve higher perceptual accuracy while transmitting fewer features compared with the Max Rate method.

Fig. 10 shows the performance of different scheduling methods under very small sensor sampling intervals in terms of AP@0.50 and AP@0.70. We consider three smaller total scheduling times of 20 ms, 50 ms, and 100 ms. It mimics the real-world situation where there is heavy congestion in the communication network or a potential driving collision, necessitating efficient scheduling during very short intervals. In this case, SchedCP-wl still demonstrates performance improvement over the baseline methods. This suggests that SchedCP-wl can obtain complementary perceptual information in a very short time, by prioritizing the scheduling of collaborators that are perceptually important and have a good channel state.

\begin{figure}[t]
\centering
\includegraphics[scale=0.4]{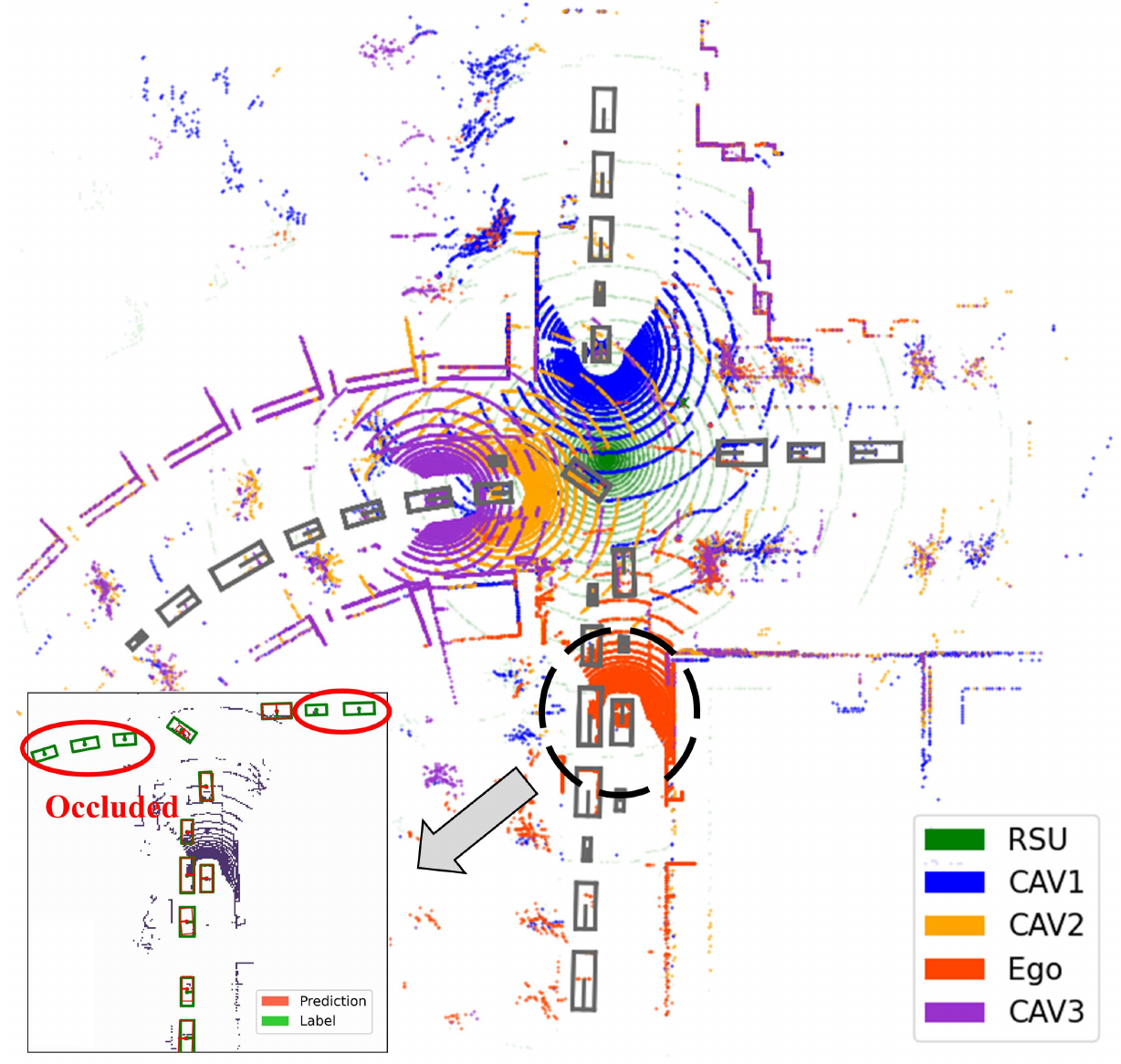}
\caption{Bird's eye view of a scenario at the intersection.}
\end{figure}

\begin{table}[t]
  \begin{center}
    \caption{Simulation Parameter}
    \setlength{\tabcolsep}{6pt}
    \vspace{-0.4cm}
    \scalebox{1.24}{
        \begin{tabular}[t]{l|c}
            \hline
            \textbf{Parameter} & \textbf{Value} \\
            \hline
            Scheduling model & SchedCP \\
            Number of collaborators & 4 \\
            Bandwidth & 200 kHz \\
            Length of scheduling slot & 5 ms \\
            Number of scheduling slots & 40 \\
            \hline
        \end{tabular}
    }
  \end{center}
\end{table}

\subsection{Case Study: An Occlusion Scenario}

\begin{figure} 
    \centering 
    \subfigure[Instantaneous transmission rate]{\label{fig:subfig:a}
    \hspace{-0.5cm} \includegraphics[scale=0.46]{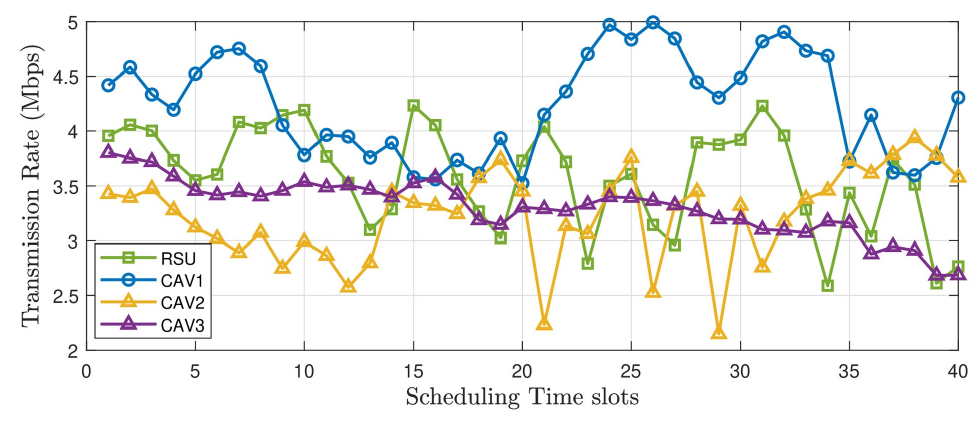}}
    \vfill
    \subfigure[Remaining confidence score]{\label{fig:subfig:a}
    \hspace{-0.5cm} \includegraphics[scale=0.46]{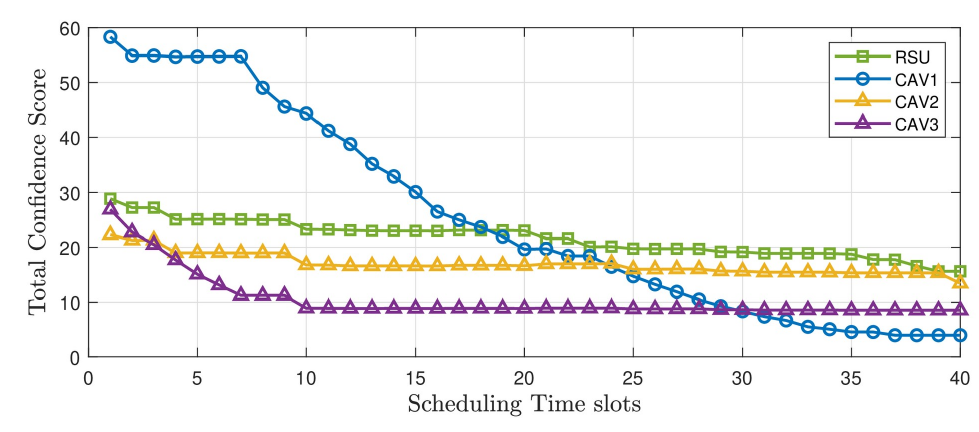}}
     \vfill
    \subfigure[Scheduling decision at every scheduling slot]{\label{fig:subfig:a}
    \hspace{-0.5cm} \includegraphics[scale=0.46]{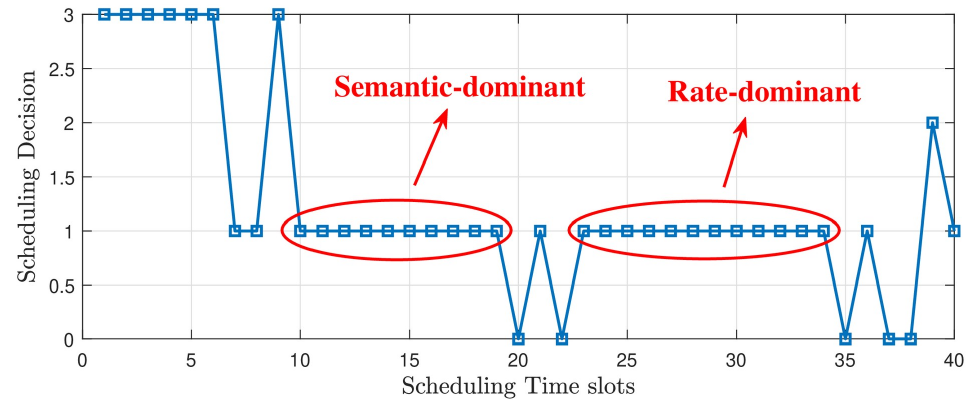}}
    \caption{Visualization of the scheduling process.}
    \label{fig:subfig}
\end{figure}

To further illustrate the advantages and mechanism of the DRL-based adaptive scheduling algorithm, we design a collaborative perception scenario with occlusion, as shown in Fig. 11. The simulation parameters are listed in Table V. As shown in Fig. 11, the ego vehicle is unable to perceive the presence of six vehicles on the front left and right sides due to the occlusion resulting from the surrounding vehicles and buildings. This can potentially lead to a vehicle collision at this intersection. Additionally, we set a relatively low bandwidth, i.e., 200 kHz, as listed in Table V, to necessitate effective V2X user scheduling.

Fig. 12 presents the vehicle scheduling process from three aspects, where Fig. 12(a) shows the variation of the instantaneous transmission rate over the sensor sampling interval, Fig. 12(b) shows the remaining confidence score possessed by each of the collaborators at each scheduling slot, which is defined by $\sum_{(x,y)}{\mathbf{R}_{j}^t}^2(x,y)$, and Fig. 12(c) shows the scheduling decisions at the beginning of each scheduling slot. In the first few scheduling slots, CAV3 is selected to access the spectrum because the three vehicles on the front left side of the ego vehicle cannot be perceived by the ego vehicle due to the occlusion, especially the vehicle behind CAV3. Other collaborators except for CAV3 cannot perceive it effectively due to far distance or occlusion while CAV3 can perceive its presence. Consequently, the scheduler selects CAV3 to access the spectrum at the beginning of the frame to transmit the BEV features associated with the target vehicle behind CAV3. 

During the period from the 10th to the 35th scheduling slot, CAV1 is predominantly selected to access the spectrum. From the 10th to the 20th slot, CAV1 has the highest total confidence score among all the collaborators as seen in Fig. 12(b), indicating that it possesses the most useful information for the perception of the ego vehicle. As shown in Fig. 11, we mention that the ego vehicle is unable to perceive the presence of the vehicles on the front left and right sides. Observing the position of CAV1 we can find that CAV1 can perceive all six vehicles well while other CAV or RSU can only perceive part of them. As a result, its perceptual features significantly improve the perception of the ego vehicle. Considering the transmission rates of different links do not vary much from the 10th to the 20th scheduling slot, CAV1 is prioritized for transmission due to its huge gain in perceptual semantics. During the period from the 20th to the 35th scheduling slots, the scheduling decision is dominantly influenced by the transmission rate of each link due to the low total confidence score of all collaborators. During this time, the transmission rate of CAV1's link with the ego vehicle is much higher than those of others, as shown in Fig. 12(a). Consequently, it is chosen to access the spectrum due to the advantage in terms of channel quality.

\begin{figure} 
    \centering 
    \subfigure[Nearest]{\label{fig:subfig:a}
    \includegraphics[width=0.44\linewidth]{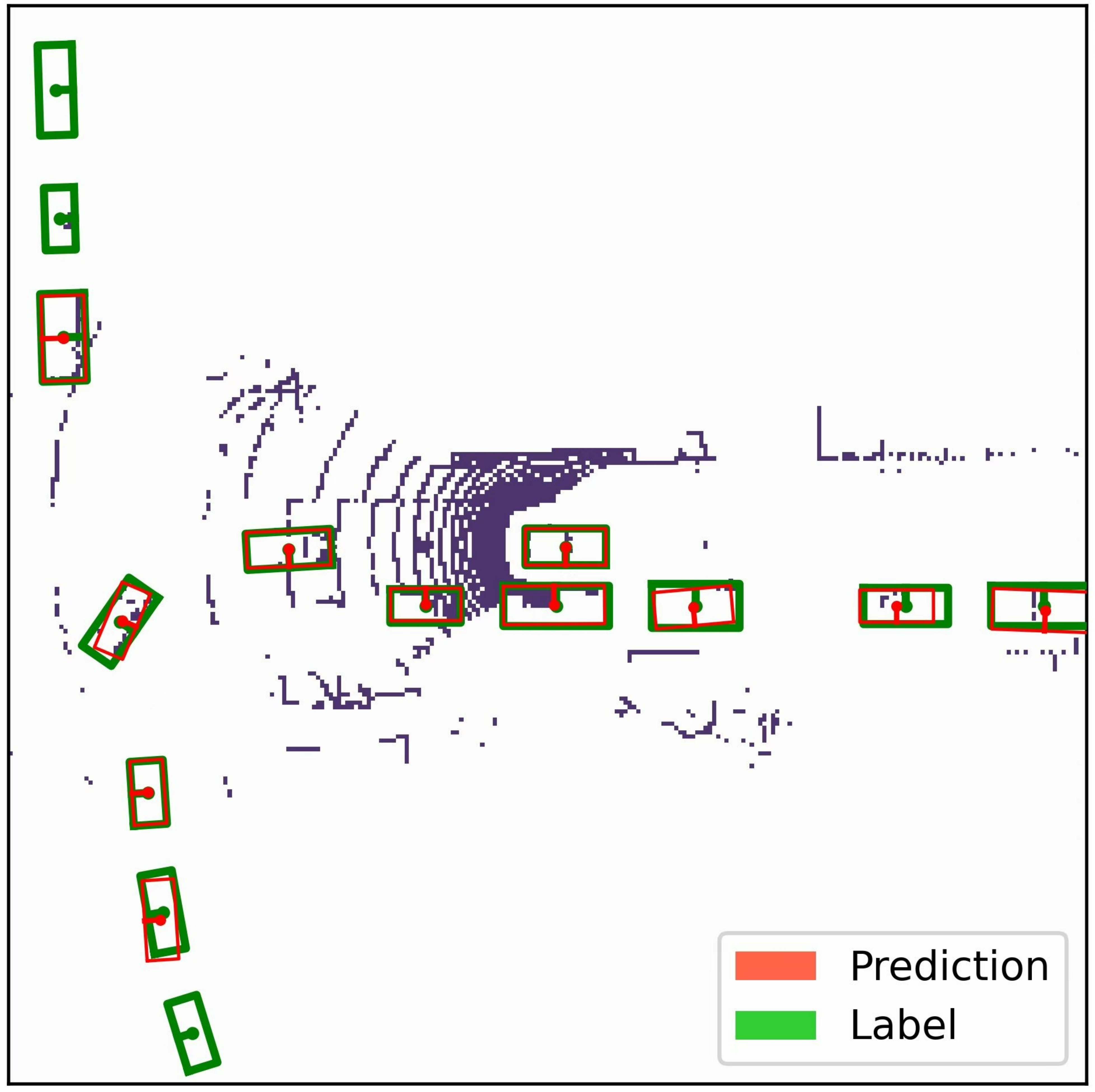}}
    \subfigure[RR]{\label{fig:subfig:a}
    \includegraphics[width=0.44\linewidth]{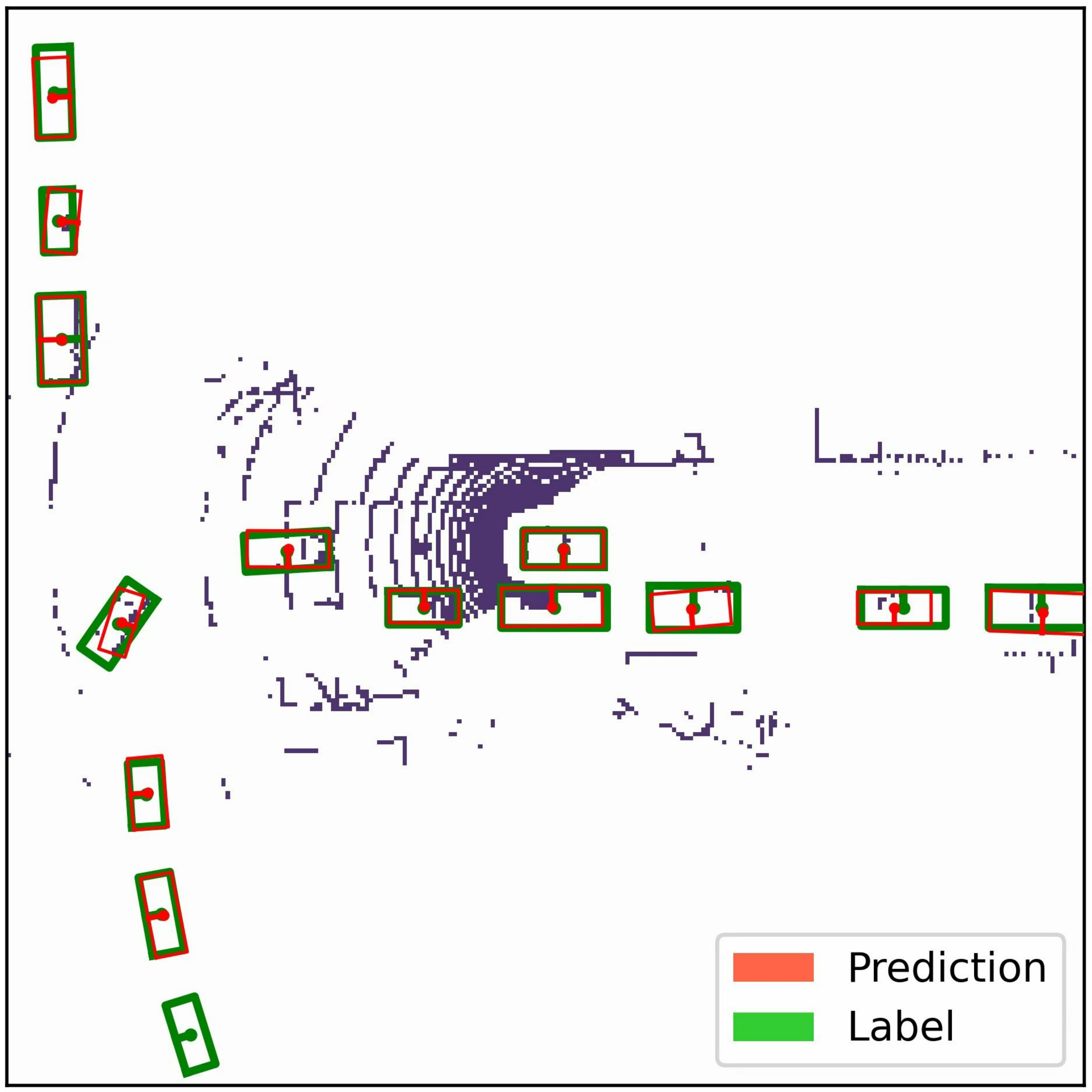}}
    \vfill
    \subfigure[Max Rate]{\label{fig:subfig:a}
    \includegraphics[width=0.44\linewidth]{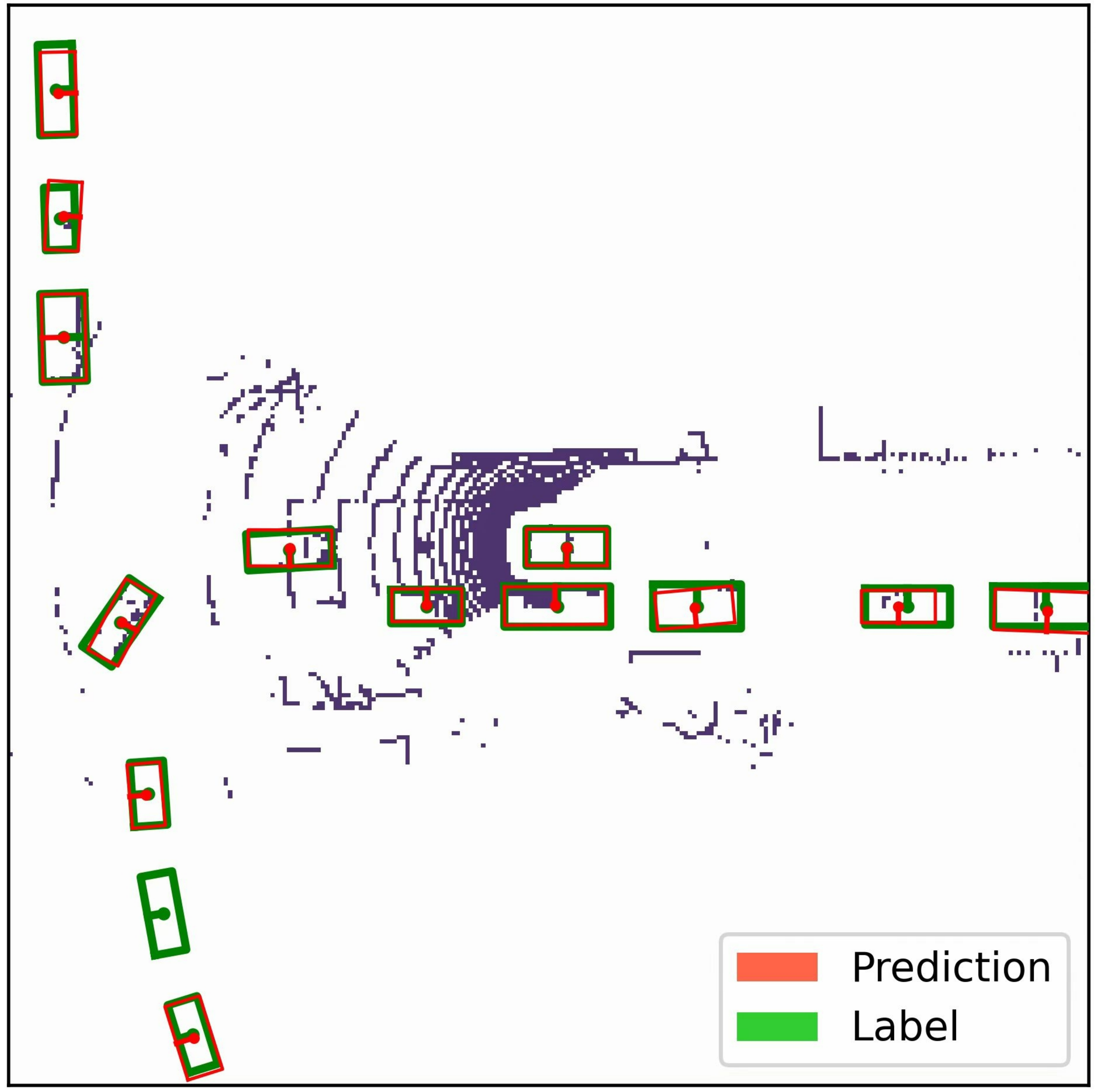}}
    \subfigure[SchedCP-wl]{\label{fig:subfig:a}
    \includegraphics[width=0.44\linewidth]{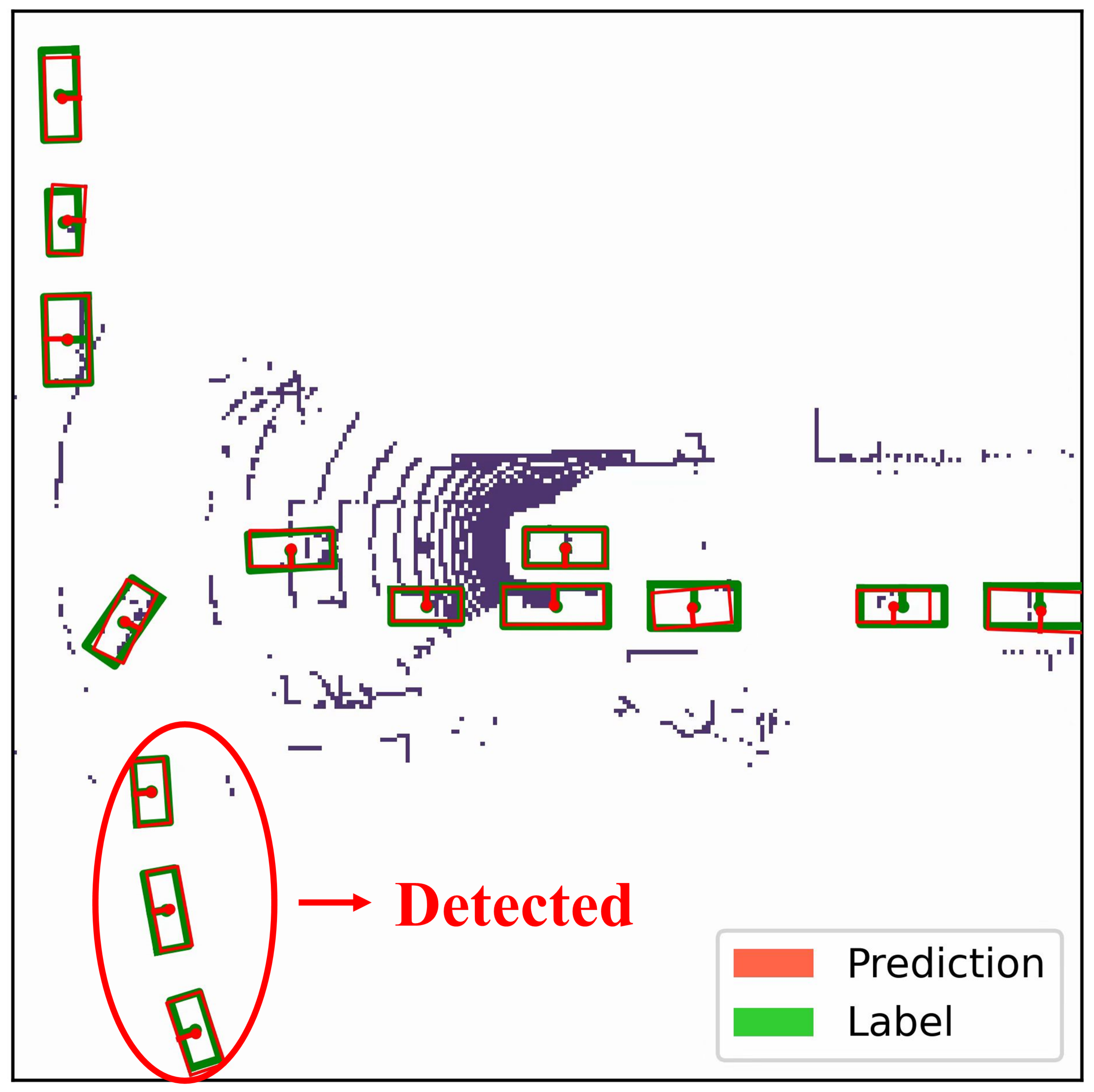}}
    \caption{Detection results of different scheduling strategies.}
    \label{fig:subfig}
\end{figure}

In addition, it is noteworthy that the scheduler selects the V2I link between the RSU and the ego vehicle at the 20th and 22nd slots. This is because, CAV1 has already shared the majority of its useful features after the previous 10 moments of transmission, which leads to its total confidence score lower than that of the RSU. And the link transmission rates of RSU and CAV1 during the 20th to 22nd scheduling slot are very similar. Combining the factor of both channel state and perceptual semantics, the V2I link between RSU and the ego vehicle is scheduled to access the spectrum.

From the analysis, it can be concluded that our proposed algorithm based on DRL can make reasonable scheduling decisions based on real-time changes in channel state and perceptual semantics. The final detection results in this scenario are shown in Fig. 13, where all three baseline methods only detect part of all target vehicles while SchedCP obtains enough perceptual information from other collaborators thus successfully detecting all target vehicles in this scenario.

\section{Conclusion}

In this paper, we propose a DRL-based V2X user scheduling algorithm for collaborative perception, named SchedCP. The cooperative-aware V2X user scheduling problem is systematically modeled with the objective of maximizing the perception performance of ego vehicles. Considering the challenge of label acquisition, we transform the original label-dependent optimization objective into a label-free optimization objective based on the characteristics of 3D object detection. Taking both channel state information and semantic information into consideration, we design a DDQN-based user scheduling framework for V2X-aided collaborative perception. We conduct extensive experiments to verify the rationality of the hypotheses and demonstrate the effectiveness of the performance of SchedCP over traditional scheduling algorithms under different bandwidth conditions. We also design a case study to illustrate how our proposed algorithm adaptively changes the scheduling decision according to real-time changes in CSI and confidence of perceptual semantics.

\small
\bibliography{IEEEabrv,conference_101719}
\bibliographystyle{IEEEtran}

\end{document}